\documentclass[10pt,twocolumn,letterpaper]{article}

\usepackage{cvpr}      

\usepackage{amsmath,amssymb}
\usepackage{booktabs}
\usepackage{multirow}
\usepackage{pifont}
\usepackage{newfloat}
\usepackage{listings}
\usepackage{tikz}
\usepackage{tabularx}
\usepackage{multirow}
\usepackage{nicematrix}
\usepackage{booktabs}
\usepackage{makecell}
\usepackage{pifont}
\usepackage{acronym}
\usepackage{subcaption}
\usepackage{soul}
\usepackage{dsfont}
\usepackage{acronym}
\usepackage{stmaryrd} 
\newcommand{\densenetbasic}{\texttt{DenseNet}$_{500}^{\text{HP}}$}
\newcommand{\densenetbasicholter}{\texttt{DenseNet}$_{500}$}
\newcommand{\ikrnetbasic}{\texttt{IKrNet}$_{500}^{\text{HP}}$}
\newcommand{\ikrnetbasicholter}{\texttt{IKrNet}$_{500}$}
\newcommand{\densenet}{\texttt{DenseNet}$_{+}^{\text{HP}}$}
\newcommand{\densenetholter}{\texttt{DenseNet}$_{+}$}
\newcommand{\ikrnet}{\texttt{IKrNet}$_{+}^{\text{HP}}$}
\newcommand{\ikrnetholter}{\texttt{IKrNet}$_{+}$}
\newcommand{\cmark}{\ding{51}}%
\newcommand{\xmark}{\ding{55}}%

\newcommand{\hr}{\text{HR}}
\newcommand{\generepolholter}{Generepol-Holter}
\newcommand{\generepol}{Generepol-Holter$_{\text{HP}}$}
\newcommand{\x}{\mathbf{x}}
\newcommand{\myeq}{\stackrel{\text{def}}{=}}
\newcommand{\baseline}{\texttt{Baseline}}
\newcommand{\nostdg}{\texttt{St-}\texttt{Dg}$_{+}$}
\newcommand{\stdg}{\texttt{St}$_{+}$\texttt{Dg}$_{+}$}
\newcommand{\suppref}[1]{Appendix~\cref{#1}}

\acrodef{ecg}[ECG]{Electrocardiogram}

\DeclareMathOperator*{\argmax}{arg\,max}

\usepackage{xcolor,colortbl}
\definecolor{Gray}{gray}{0.85}
\newcolumntype{a}{>{\columncolor{Gray}}c}
\definecolor{cvprblue}{rgb}{0.21,0.49,0.74}
\usepackage[pagebackref,breaklinks,colorlinks,allcolors=cvprblue]{hyperref}

\def\paperID{13576} 
\def\confName{CVPR}
\def\confYear{2025}

\title{IKrNet: A Neural Network for Detecting Specific Drug-Induced Patterns in Electrocardiograms Amidst Physiological Variability}

\author{
Ahmad Fall$^{1}$ \quad
Federica Granese$^{1}$ \quad
Alex Lence$^{1}$ \\
$^{1}$IRD, Sorbonne Université,
Unité de Modélisation Mathématique et Informatique \\ des Systèmes Complexes, UMMISCO
F-93143 Bondy, France \\
{\tt\small \{ahmad.fall, federica.granese, alex.lence\}@ird.fr}
\and
Dominique Fourer$^{2}$ \quad
Blaise Hanczar$^{2}$ \\
$^{2}$University of Evry - Paris-Saclay, IBISC Laboratory
Evry, France \\
{\tt\small \{dominique.fourer, blaise.hanczar\}@univ-evry.fr}
\and
Joe-Elie Salem$^{3,4}$ \\
$^{3}$Clinical Investigation Center Paris-Est, CIC-1901, INSERM, Department of Pharmacology \\ Pitié-Salpêtrière University Hospital, Sorbonne Université, France \\
$^{4}$Department of Medicine, Vanderbilt University Medical Center \\ Nashville, TN, USA \\
{\tt\small joe-elie.salem@aphp.fr}
\and
Jean-Daniel Zucker$^{1,5}$ \quad
Edi Prifti$^{1,5}$ \\
$^{5}$Sorbonne Université, INSERM, Nutrition et Obesities; systemic approaches, \\
NutriOmique, AP-HP Hôpital Pitié-Salpêtrière, France \\
{\tt\small \{jean-daniel.zucker, edi.prifti\}@ird.fr}
}

\begin{document}
\maketitle
\begin{abstract}
Monitoring and analyzing electrocardiogram (ECG) signals, even under varying physiological conditions, including those influenced by physical activity, drugs and stress, is crucial to accurately assess cardiac health. However, current AI-based methods often fail to account for how these factors interact and alter ECG patterns, ultimately limiting their applicability in real-world settings. This study introduces IKrNet, a novel neural network model, which identifies drug-specific patterns in ECGs admist certain physiological conditions. 
IKrNet's architecture incorporates spatial and temporal dynamics by using a convolutional backbone with varying receptive field size to capture spatial features. A bi-directional Long Short-Term Memory module is also employed to model temporal dependencies. 
By treating heart rate variability as a surrogate for physiological fluctuations, we evaluated IKrNet’s performance across diverse scenarios, including conditions with physical stress, drug intake alone, and a baseline without drug presence. Our assessment follows a clinical protocol in which 990 healthy volunteers were administered 80mg of Sotalol, a drug which is known to be a precursor to Torsades-de-Pointes, a life-threatening arrhythmia. We show that IKrNet outperforms state-of-the-art models' accuracy and stability in varying physiological conditions, underscoring its clinical viability.
\end{abstract}   
\section{Introduction}
\label{sec:introduction}
An \acf{ecg} is a diagnostic tool that records the heart’s electrical activity through electrodes placed on the body, each producing an ECG lead. ECG leads provide multiple spatial and temporal perspectives on cardiac activities, aiding in diagnosing heart conditions.

In traditional clinical settings, ECGs are recorded over short, 10-second intervals while the patient rests. These recordings are commonly taken during in-clinic medical examinations and effectively detect persistent cardiac conditions. However, they may fail to capture transient or intermittent abnormalities outside controlled environments. For instance, long-term ECG monitoring is crucial for detecting life-threatening arrhythmias, such as Torsades-de-Pointes (TdP)~\cite{viskin1999long, yap2003drug} — a specific form of polymorphic ventricular tachycardia marked by rapid, irregular heartbeats that can lead to sudden cardiac arrest.

To overcome this limitation, wearable devices such as Holter monitors facilitate continuous ECG monitoring over 24-48 hours or longer, allowing one to observe real-world cardiac activity as patients go about their daily lives. However, continuous monitoring introduces additional challenges. Holter devices typically operate at a lower frequency range (180-250 Hz)~\cite{Winokur2013-ov}, use fewer leads than the usual 12 used in clinical settings, and are susceptible to interference from factors like physical activity, stress, drugs and dietary influences~\cite{bravi2013physiological}. These factors can affect the cardiovascular system, introducing distinct but complex ECG patterns that are further complicated by high noise levels and physiological variability over extended periods.

Manual analyses of Holter recordings are time-intensive, often requiring disentangling overlapping signal patterns. This challenge is exacerbated by heart rate variability (HRV), which is sensitive to physiological and pharmacological influences~\cite{LI2023108128}. Our study assesses TdP risk by using Sotalol~\cite{HOHNLOSER1993A67} as a pharmacological surrogate to induce QT prolongation, creating conditions that mimic the arrhythmogenic environment leading to TdP. To bridge the clinical challenges of detecting TdP in real-world settings with advancements in deep learning (DL), we focus on using DL models to improve the reliability of ECG analysis under varying physiological conditions. 

DL has shown promise in ECG applications, such as noise reduction~\cite{7530192,8693790}, waveform segmentation~\cite{moskalenko2019deep}, feature extraction~\cite{alam2023qtnet}, and abnormality detection~\cite{ansari2023deep}. Most architectures, including ResNet~\cite{he2015deepresiduallearningimage, 10.3389/fphys.2024.1362185}, Long Short-Term Memory (LSTM)~\cite{10.1162/neco.1997.9.8.1735, roy2023ecg}, and Transformer~\cite{zhao2023transformingecgdiagnosisanindepth, akan2023ecgformer}, have shown potential for automated ECG analyses and arrhythmia detection. With their residual connections, ResNet models are adept at capturing spatial patterns in ECG signals, while LSTMs are well-suited for capturing temporal dependencies in sequential data like ECG waveforms. Transformer models, on the other side, leverage self-attention mechanisms~\cite{vaswani2023attentionneed} to model long-range dependencies in sequential data, which is beneficial for ECG analysis.

However, these models face significant limitations when applied in real-world scenarios, particularly in the presence of substantial HRV. ResNet’s focus on spatial features limits its effectiveness in continuous temporal tracking, which is crucial for detecting subtle QT prolongations and arrhythmogenic patterns influenced by HRV. For instance, LSTMs struggle to adapt to complex interactions between physiological HRV and drug-induced effects, making them less robust under abrupt changes in ECG signals, as they do not have internal mechanisms to focus on specific parts of the waveform. Transformers, while capable of modeling global dependencies, require extensive data and are prone to noise and variability in ambulatory ECGs — especially at lower sampling rates typical of Holter monitors. Additionally, the computational demands of Transformer models often preclude their real-time or on-device application due to limited resources~\cite{zhao2023transformingecgdiagnosisanindepth}.
Previous studies, such as~\cite{prifti2021deep}, have developed CNN-based models for predicting the presence of Sotalol as a surrogate for assessing drug-induced long QT syndromes (diLQTS) and congenital long QT syndromes (cLQTS), both of which are risk factors for Torsades-de-Pointes. Their most effective model, a DenseNet~\cite{huang2018denselyconnectedconvolutionalnetworks} with six blocks of dense convolutional layers, improved diLQTS and cLQTS detection but in expert hand-picked clean ECGs. However, the authors did not evaluate the impact of HRV on the model’s performance, leaving a critical gap in robustness for real-world applications.
To address these limitations, our study advances the field by making the following contributions:\\
\noindent\textbf{1. We formally describe our proposed framework}, designed to be flexible enough for application across various contexts. Additionally, we establish a robustness definition tailored to our needs, introducing a fairness-inspired metric that quantitatively assesses the model’s robustness.\\
\noindent\textbf{2. We present a new architecture, \texttt{IKrNet}, specifically designed to detect drug-induced ECG patterns}. Our model leverages both spatial and temporal components of ECG signals through a multi-branch structure, using varying receptive field size to capture features across multiple scales. Spatial features are extracted, followed by bi-directional LSTM layers capturing temporal dependencies.\\
\noindent\textbf{3.} We evaluate both state-of-the-art models and our proposed model on a clinician-curated dataset and a real-world dataset obtained from Holter monitors. \textbf{This comparison reveals the limitations of traditional models in handling data with physiological variability and different sampling rates. Additionally, we rigorously assess model performance by considering typically overlooked ECG features}, such as heart rate variability, which can significantly influence drug-related ECG patterns.

\section{Framework} 
\label{sec:preliminaries}
We introduce a new classifier, named IKrNet, designed to detect \textit{drug footprints} from ECGs robustly. By \textbf{detecting drug footprint}, we refer to the classifier's ability to identify unique patterns in ECG signals induced explicitly by the drug. By \textbf{robustly}, we mean that the classifier accurately classifies ECGs despite variations in interactions from other footprints, such as stress, physical activity, or natural physiological fluctuations.

\begin{figure}[h!]
    \centering
    \includegraphics[width=\columnwidth]{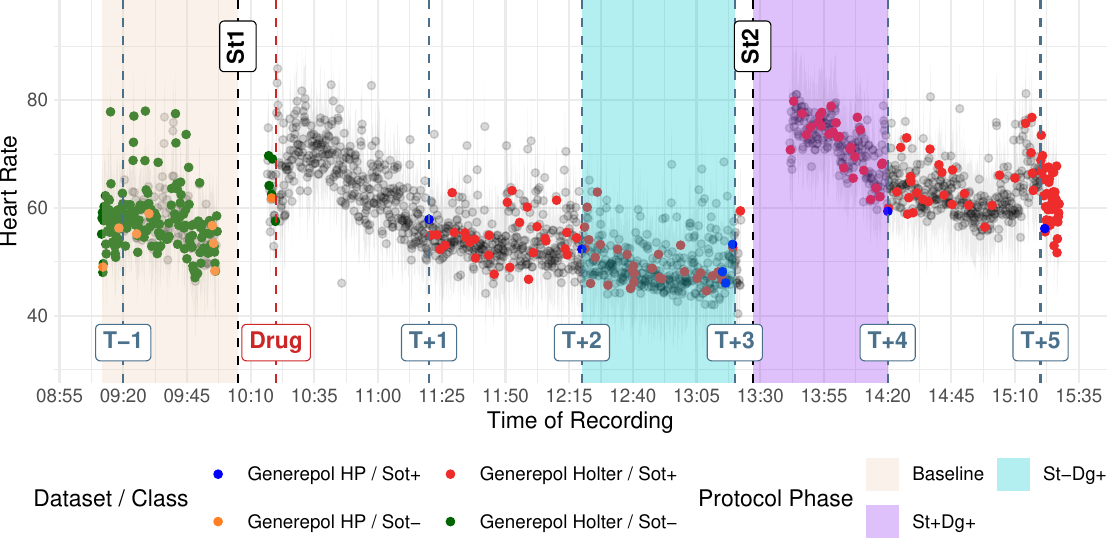}
    \caption{\textbf{Clinical protocol setup (the GENEREPOL study):} The heart rate is depicted for each of the 10s ECGs, recorded with a Holter device for one patient. Each ECG is depicted by a dot, which is coloured in grey when not used during the training, in green and blue respectively before drug intake (Sot-) and after drug intake (Sot+) for the Holter dataset and in orange and blue for the expert hand-picked dataset. Because of the unbalanced ECGs before and after drug intake, we selected a proportionate amount of ECGs for the training process (cf.~\cref{sec:dataset})}
    \label{fig:protocol}
\end{figure} 

We focused on the use case described in~\cite{salem2017genome}, where data from 990 healthy participants from the GENEREPOL study received 80mg of Sotalol, an IKr blocker. This drug inhibits the rapid component of the delayed rectifier potassium current (IKr), which is crucial for cardiac repolarization and leads to prolonged QT intervals on the ECG. This prolonged repolarization phase increases the risk of Torsade-de-Pointes. The IKr footprint refers in part to this QT prolongation, although other more subtle alterations are suspected and serves as a surrogate marker for TdP risk. Indeed, actual TdP events are rare and it is difficult to capture the risk before such events in a study setting, especially when vast datasets are needed to train DL models. Study participants were fitted with portable Holter monitors to record their hearts' electrical activity in real time before, during, and after Sotalol administration. As illustrated in~\cref{fig:protocol}, ECGs were recorded continuously over six hours, starting one or two hours before (here denoted as baseline) and continuing for up to five hours after Sotalol intake. Before taking the drug, participants underwent a stress test; they completed a second stress test three hours after taking the drug.

This section formalizes our framework, for broad applicability. We define key concepts and introduce the drug-footprint task as a supervised learning problem, along with the metrics used to evaluate the model's robustness. 

\subsection{Preliminaries}
\label{sec:preliminaries}
We consider the case of a single-lead \ac{ecg} signal $x$ (each observation of the heart activity is referred to as a lead). Here, $x$ can be viewed as a sequence of $K$ time-varying impulsive patterns $x_i$ delimited by consecutive time instants $t_{i-1}$, $t_{i}$ expressed in seconds, with $t_{i-1} < t_i$, such as:

\begin{equation}
x(t) = \begin{cases} x_1(t) & \text{if~} t \in [t_0, t_1[\\ 
x_2(t) & \text{if~} t \in [t_1, t_2[\\
 & \vdots \\
x_K(t) & \text{if~} t \in [t_{K-1}, t_{K}]\\
\end{cases} 
\end{equation}

Each component $x_i$ corresponds to a heart beat. 
\paragraph{Discretization/Sampling process.}
Given a continuous-time signal $x(t)$, we 
can obtain its discretized version at sampling
frequency $F_s$ expressed in Hz, assuming a uniform sampling 
process as follows:
\begin{equation}
x[n] = \displaystyle\int_\mathbb{R} x(t) \delta(t-\frac{n}{F_s}) \mathrm dt = x(\frac{n}{F_s})
\end{equation}
where $\delta$ is the Dirac distribution, $n \in \llbracket0 , N-1\rrbracket$ 
denotes the time index, where $N = \lfloor t_K F_s \rfloor$, is the fine-length of the input vector $\x = \left[x[0], x[1],\cdots, x[N-1] \right]$. According to the Nyquist-Shannon theorem, such signal enables a perfect reconstruction if $F_s \geq 2f_{\max}$, where $f_{\max}$ is the maximal frequency present in the signal. 

\subsection{Drug-footprint detection task}
\label{sec:framework}
Let $X$ be the random variable (r.v.) for the ECG recordings, with realizations $\mathbf{x} \in \mathcal{X} \subseteq \mathbb{R}^N$, where $\mathcal{X}$ is the space of ECG as defined in~\cref{sec:preliminaries}. Let $Y$ be the r.v. representing the class label, taking values in $\mathcal{Y} = \{0, 1\}$, where 1 indicates the presence of the drug footprint, and 0 indicates its absence. The samples $(\mathbf{x}, y)$ are i.i.d. from $P_{XY}$.

The classifier $f_\theta : \mathcal{X} \rightarrow \mathcal{Y}$, parameterized by $\theta$, maps an input ECG to a label, estimating whether the recording shows signs of the drug effect. We define this classifier as
$
f_\theta(\x) = \arg\max_{y \in \mathcal{Y}} p_{\widehat{Y}|X}(y|\x, \theta),
$
where $p_{\widehat{Y}|X}(\cdot|\x, \theta)$ is the model’s posterior distribution. In~\cref{sec:ikrnet}, we describe our proposed architecture tailored for drug-footprint detection and the rationale behind its design choices.

\subsection{Evaluating model robustness}
\label{sec:robustness}
Finally, assuming that $f_\theta$ is trained without awareness of subgroup distinctions, we define a fairness-inspired notion of robustness to evaluate model stability across diverse recording conditions. Specifically, we frame this as an accuracy parity problem~\cite{barocas2023fairness}, where $f_\theta$ is considered robust across the subgroups $\mathcal{X}^{(1)}, \dots, \mathcal{X}^{(m)}$, which are subsets of $\mathcal{X}$, if for any subgroups $\mathcal{X}^{(i)}$ and $\mathcal{X}^{(j)}$:
\[
\left|\mathbb{P}(Y = f_\theta(X^{i}) \mid X^{i}) - \mathbb{P}(Y = f_\theta(X^{j}) \mid X^{j})\right| = 0,
\]
where $X^{\bullet}$ is the r.v. whose realizations are in $\mathcal{X}^{\bullet}$ and $Y$ is the true label. 

\paragraph{Practical implementation.} In our clinical use case, we consider the protocol time for identifying the subgroups we would like the model to achieve similar performances.
By considering the protocol in~\cref{fig:protocol}, we recognise the following subgroups (or zones)\footnote{The \texttt{Stress} zone, which includes ECGs with a high heart rate and no drug administration, is omitted due to the lack of available data.}, (i) {\baseline} – from the start of the recording to the first stress test (\textit{St1}); (ii) No Stress and Drug Present (\nostdg) – from \textit{T+2h} to \textit{T+3h}, where no stress test is conducted but the drug is active; (iii) Stress and Drug Present (\stdg) – immediately after the second stress test, where both the stress test and drug are present.
In this context, we will call such subgroups \underline{protocol zones}.
We say $f_\theta$ is robust w.r.t. 
the protocol zones if\footnote{The optimal scenario is identical accuracy across protocol zones. In practice, we consider a model more robust if this difference is \textit{reasonably small}.}
$$
\left |\,acc_{\max} - acc_{\min}\, \right|
 = 0,
$$
where 
$acc_{\text{func}}\myeq\text{func}\{acc(\baseline), \allowbreak acc(\texttt{St-}\texttt{Dg}_{+}), acc(\texttt{St}_{+}\texttt{Dg}_{+})\}$ with $\text{func}\in\{\min, \max\}$, and 
$acc(\mathcal{S})\myeq \frac{1}{|\mathcal{S}|}\sum_{(\x, y)\in\mathcal{S}}\mathds{1}\left[f_\theta(\x)=y\right]$.

\section{Experimental Setting}
\label{sec:experimental_setting}
In the following section, we provide details of the experimental setup. In particular,~\cref{sec:ikrnet} contains a thorough description of the proposed architecture.

\subsection{Datasets and preprocessing}
\label{sec:dataset}
We used the {\generepolholter} and {\generepol} datasets for our experiments. The {\generepolholter} dataset contains all the 10s consecutive ECGs from 990 participants recorded at a sampling rate of 500 Hz, following the clinical protocol presented in~\cref{fig:protocol} and  in~\cite{salem2017genome}.  The {\generepol} dataset comprises a total of 15119 ECGs (~15/participant) manually selected by expert cardiologists from the {\generepolholter} dataset. These hand picked (HP) ECGs are relatively clean \textit{(in the clinical sense)} and contain a low noise level. However, they do not accurately reflect real-life scenarios, as most recorded ECGs are subject to different perturbations, including noise.
Both datasets label ECGs as Sot- (before Sotalol intake) or Sot+ (after) and are divided into  training (70\%), validation (10\%), evaluation (10\%, to pick the best hyper-parameters configuration), and holdout (10\%) partitions. To prevent class imbalance in {\generepolholter}, an equal number of Sot+ and Sot- ECGs were randomly selected per study participant. Indeed, Sot- ECGs are only recorded for a maximum of 2 hours, whereas Sot+ ECGs are recorded for 5 hours. Each participant had approximately the same amount of selected ECGs in the different protocol times (T+1, T+2, T+3, T+4, T+5 in~\cref{fig:protocol}) for label Sot+. The {\generepol} training partition contained 6315 ECGs, and 61544 ECGs for the {\generepolholter} dataset.

\paragraph{Data augmentation techniques.}
To enhance data variability and address the varying sampling rates of recording devices, which typically operate at $F_s \in {180, 250, 500, 1000}$ Hz, we applied a sampling rate augmentation technique~\cite{9983696, Kwon2018-xc}. This process involves downsampling an original ECG signal (sampled at 500 Hz) to a lower rate, $F^\prime_s \in {180, 250}$ Hz, and then upsampling it back to 500 Hz for the training, validation, and evaluation sets.
This simulates ECG recordings initially captured at lower sampling rates and then upsampled, reflecting real-world variability in ECG data~\cite{BAUMERT2016159}. We used spline interpolation for both downsampling and upsampling, in line with standard ECG processing techniques~\cite{SIDEK201313}.
After data augmentation, the {\generepol} training partition contained 18945 ECGs, and the {\generepolholter} dataset 184631 ECGs. Detailed number of ECGs per partition is provided in~\suppref{tab:generepol_ecg_distribution}.

\paragraph{Data preprocessing.}
We standardized the ECG signals before inputting them into the network. This process aligns input feature scales, which is essential to prevent the model from applying disproportionate weight to ECGs with higher amplitudes~\cite{SHANKER1996385}. By doing so, we mitigate the influence of amplitude variations that may arise from factors unrelated to the underlying cardiac conditions, such as differences in recording equipment, or individual patient characteristics. 
\subsection{{\texttt{IKrNet}}: our novel proposed architecture}
\label{sec:ikrnet}
\begin{figure}[h]
    \centering
    \includegraphics[width=\columnwidth]{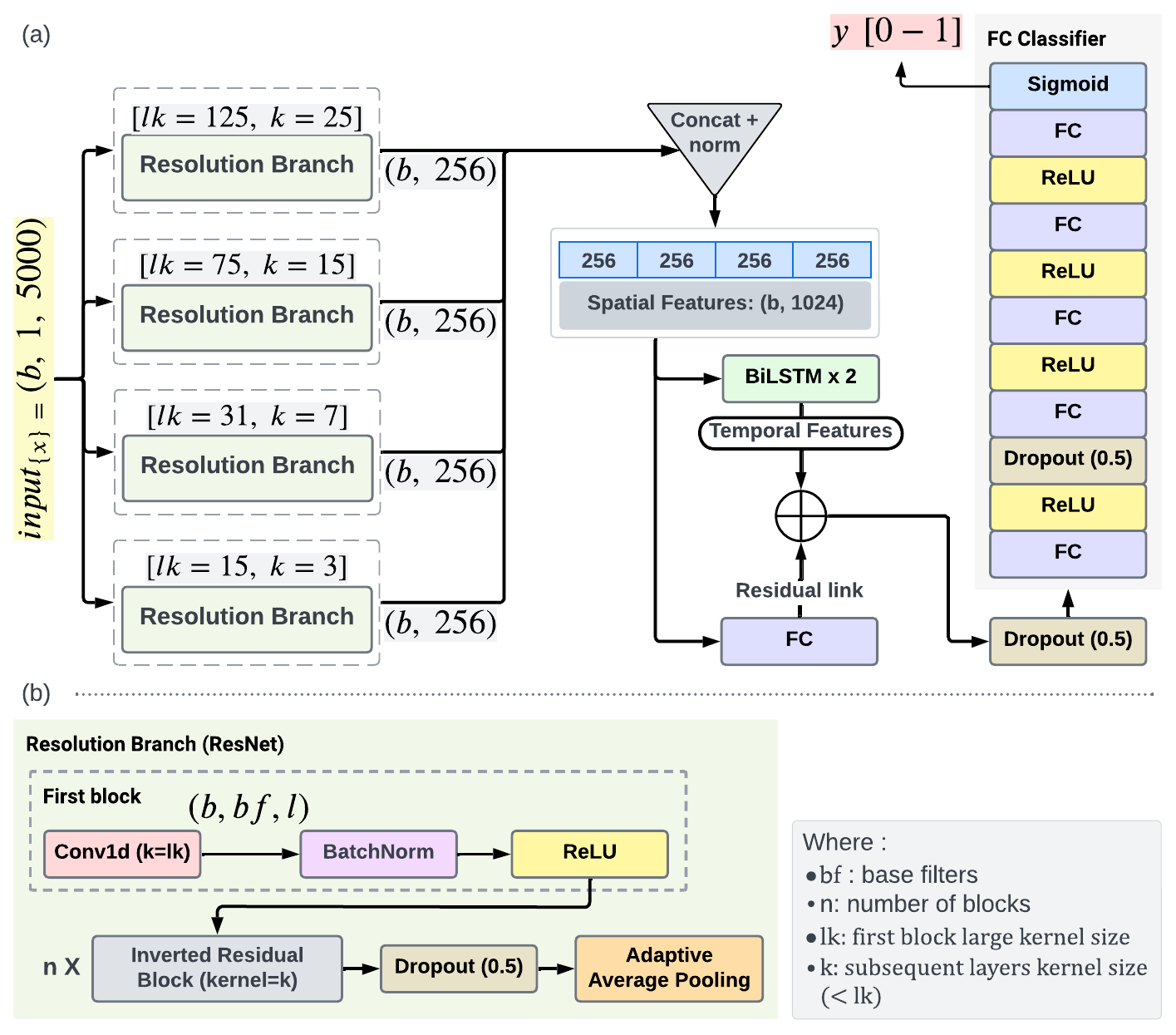}
    \caption{(a) Global architecture of \texttt{IKrNet}, featuring a CNN backbone with multiple branches, each capturing low- and high-frequency features through different receptive fields. (b) Detailed branch structure with inverted residual blocks, optimizing feature extraction and reducing complexity.}
    \label{fig:ikrnet_model_architecture}
\end{figure}
In~\cref{fig:ikrnet_model_architecture}, we illustrate the \texttt{IKrNet} architecture. A CNN backbone~\cite{lecun_deep_2015} first extracts spatial patterns from the raw ECG signal, using multiple branches with varying receptive field widths to capture features across different scales. After spatial feature extraction, bi-directional Long Short-Term Memory (BiLSTM) layers~\cite{Cheng2021-hl, huang2015bidirectionallstmcrfmodelssequence} analyze temporal dependencies in the ECG signal. Both spatial and temporal features are summed to produce a consolidated feature vector of the ECG, then fed to a fully connected classifier, which leverages the combined feature set to predict the occurrence of the drug footprint.

\paragraph{Spatial feature extractor.}
It comprises multiple resolution branches that process the same signal input, each with a distinct receptive field, enabling a multi-resolution analysis that adapts to varying ECG sampling rates and captures both low- and high-frequency variations. This design enhances the model's ability to detect subtle and pronounced drug-induced changes, such as QT prolongation and heart rate changes associated with the drug. Within each branch, we employ a ResNet~\cite{he2015deepresiduallearningimage, 10.3389/fphys.2024.1362185} module known for its ability to facilitate the training of deeper network structures through identity shortcut connections~\cite{he2015deepresiduallearningimage}. These connections help mitigate the vanishing gradient problem by allowing gradients to flow directly through the network layers. 

Each ResNet begins with a convolution layer with a large initial kernel size $(lk)$, designed to capture macro features. This is followed by a series of residual blocks that employ a smaller, consistent kernel size $(k)$, extracting more profound and detailed patterns. The distinct kernel pair $(lk, k)$ varies across each branch, enabling the architecture to achieve multi-resolution analysis effectively. A larger kernel $(lk)$ captures information over longer distances, while a smaller kernel $(k)$ focuses on nearby information. 
Instead of basic blocks~\cite{he2015deepresiduallearningimage} as in the original ResNet architecture, we used inverted residual structures~\cite{sandler2019mobilenetv2invertedresidualslinear} as described in~\suppref{fig:model_architecture_inverted_residual_block}, which enhance model efficiency. 
They significantly reduce the trainable parameters, thus reducing computational costs compared to traditional basic blocks by limiting the scaling of parameters with the kernel size and number of channels. Squeeze-and-excite layers~\cite{8578843} (see~\suppref{fig:model_architecture_inverted_squeeze_and_excite}), are integrated to recalibrate channels dynamically, compressing feature maps via global average pooling and modulating them with a fully connected layer and sigmoid activation. This process enhances sensitivity to clinically relevant ECG patterns. Features from each branch are then pooled with adaptive average pooling to reduce dimensionality and consolidate key information for temporal analysis. Initial inputs undergo progressive downsampling through convolution strides.

\paragraph{Temporal feature extractor.}
After extracting spatial features, the model processes temporal features to learn the dynamic nature of ECGs. This requires recognizing static patterns and capturing their temporal evolution to predict the drug intake footprint. Two BiLSTM layers process the data bidirectionally, capturing dependencies across the sequence to characterize cardiac events. Each BiLSTM layer integrates the concatenated spatial features extracted by the CNNs, building a complete temporal profile of the ECG. 

\paragraph{Drug footprint detection.}
At this stage, we combine the temporal and the spatial features, which involves a skip connection strategy that directly adds the output from the fully connected layer to the output of the last BiLSTM layer. This approach preserves both high-resolution spatial details and deep temporal relationships. The final integration of features is then fed into a fully connected classifier that produces a score between 0 and 1.

\paragraph{Training process}
\label{sec:ikr_training}
We trained our model using the binary cross-entropy loss function alongside the AdamW optimizer~\cite{loshchilov2019decoupledweightdecayregularization}, an improved version of the traditional Adam optimizer, introducing decoupled weight decay regularization allowing better regularization, and reduced overfitting by penalizing large weights. The training was conducted over ten epochs, using 4 NVIDIA A100 GPUs, each with 80 GB of memory, in approximately 4 hours. The model exhibited rapid convergence throughout the training, with both training and validation losses decreasing steadily and remaining closely aligned, suggesting the model does not overfit the training data.
A hyperparameter optimization process and an ablation study (cf.~\cref{sec:ablation_study}) were conducted to identify the optimal hyperparameter set. The best-performing model, selected based on evaluation metrics, had the following hyperparameters: batch size of 64 per GPU (256 globally), a learning rate of 0.001, 4 branches in the CNN backbone with kernel pairs ($lk$, $k$) $(125,25), (75,15), (31,7), (15,3)$. Each CNN branch had the following strides $[1, 5, 1, 5, 1, 4, 1, 4, 1, 3]$, an initial convolution filter of $64$, and $10$ inverted residual blocks. The convolution filters were increased by $2$ after every $4$ block. The final architecture has $\sim$24 million parameters.

\section{Experimental Results and Analysis}
\label{sec:results}

\begin{table*}[h]
    \centering
    \caption{\textbf{Cross-dataset model accuracies and Accuracy Parity Difference (APD) by protocol zone across sampling rates:} The results are presented as mean$\pm$standard deviation for ECGs sampled at 150, 180, 215, 250, 300, 350, 425, and 500 Hz.}
    \resizebox{2\columnwidth}{!}{
    \begin{tabular}{l|c|c|c|a||c|c|c|a}
    \toprule
    & \multicolumn{4}{c||}{\textbf{\generepol}} & \multicolumn{4}{c}{\textbf{\generepolholter}} \\
    \cmidrule{2-9}
    \multirow{3}{*}{} & 
    \multicolumn{1}{c|}{{\baseline}} & \multicolumn{1}{c|}{{\nostdg}} & 
    \multicolumn{1}{c|}{{\stdg}} & \multicolumn{1}{a||}{\textbf{-}} & 
    \multicolumn{1}{c|}{{\baseline}} & \multicolumn{1}{c|}{{\nostdg}} & 
    \multicolumn{1}{c|}{{\stdg}} & \multicolumn{1}{a}{\textbf{-}}\\
    \cmidrule{2-9}
        & Accuracy ($\uparrow$)  & Accuracy ($\uparrow$) & Accuracy ($\uparrow$)& APD ({\footnotesize $\downarrow_{0}$}) & Accuracy ($\uparrow$) & Accuracy ($\uparrow$) & Accuracy ($\uparrow$)& APD ({\footnotesize $\downarrow_{0}$})\\
          \midrule
          \multicolumn{9}{l}{\textbf{Models trained on {\generepol}}} \\
          \midrule 
\densenetbasic   &  88.82 $\pm$ 8.98 & 95.08 $\pm$ 2.68 & 98.61 $\pm$ 0.41 & 11.71 $\pm$ 10.18 & 49.44 $\pm$ 14.27 & \textbf{99.64} $\pm$ 0.19 & \textbf{96.96} $\pm$ 1.41 & 59.49 $\pm$ 19.74 \\
\densenet   &  91.64 $\pm$ 5.27 & 94.15 $\pm$ 2.01 & \textbf{98.92} $\pm$ 0.7 & 8.7 $\pm$ 5.96 & 65.13 $\pm$ 12.83 & 99.59 $\pm$ 0.15 & 93.98 $\pm$ 1.49 & 39.37 $\pm$ 16.84 \\
\ikrnetbasic   &  92.87 $\pm$ 3.86 & \textbf{98.0} $\pm$ 0.37 & 97.65 $\pm$ 0.66 & 5.68 $\pm$ 4.38 & 78.43 $\pm$ 6.43 & 99.2 $\pm$ 0.12 & 92.97 $\pm$ 0.89 & 22.75 $\pm$ 7.58 \\
\ikrnet   &  94.92 $\pm$ 0.92 & 94.21 $\pm$ 1.54 & 98.37 $\pm$ 0.21 & 4.53 $\pm$ 1.59 & 79.89 $\pm$ 2.08 & 99.08 $\pm$ 0.13 & 92.31 $\pm$ 0.59 & 20.91 $\pm$ 2.33 \\
          \midrule
          \multicolumn{9}{l}{\textbf{Models trained on {\generepolholter}}} \\
          \midrule
\densenetbasicholter   &  94.73 $\pm$ 1.43 & 90.23 $\pm$ 1.36 & 94.9 $\pm$ 0.78 & 5.52 $\pm$ 1.31 & 85.68 $\pm$ 2.05 & 96.39 $\pm$ 0.63 & 86.23 $\pm$ 0.56 & 12.67 $\pm$ 1.28 \\
\densenetholter   &  96.28 $\pm$ 0.69 & 89.9 $\pm$ 1.1 & 95.42 $\pm$ 0.54 & 6.97 $\pm$ 1.21 & 89.74 $\pm$ 1.85 & 97.17 $\pm$ 0.38 & 89.37 $\pm$ 0.42 & 9.31 $\pm$ 0.73 \\
\ikrnetbasicholter   &  97.57 $\pm$ 1.6 & 91.89 $\pm$ 0.25 & 97.21 $\pm$ 0.65 & 6.52 $\pm$ 0.53 & 89.69 $\pm$ 4.07 & 99.1 $\pm$ 0.12 & 96.08 $\pm$ 0.61 & 9.88 $\pm$ 4.45 \\
\ikrnetholter   &  \textbf{97.72} $\pm$ 1.28 & 95.48 $\pm$ 0.4 & 98.65 $\pm$ 0.41 & \textbf{3.39} $\pm$ 0.7 & \textbf{92.22} $\pm$ 1.66 & 98.92 $\pm$ 0.19 & 95.35 $\pm$ 0.27 & \textbf{6.98} $\pm$ 1.67 \\
    \bottomrule    
    \end{tabular}
    }
    \label{tab:results_steps_all}
\end{table*}

\begin{figure*}[h]
  \centering
  \begin{minipage}{0.48\linewidth}
    \begin{subfigure}{1.0\linewidth}
      \includegraphics[width=\linewidth]{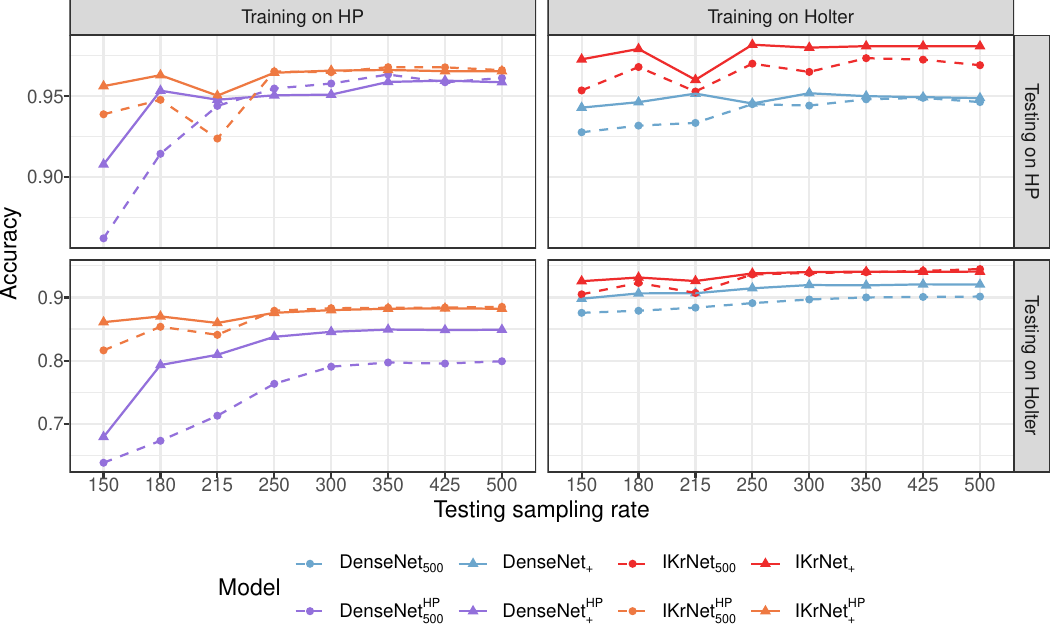}
      \subcaption{}
      \label{fig:accuracy_per_sr_all_datasets}
    \end{subfigure}
  \end{minipage}
  \hfill
  \begin{minipage}{0.48\linewidth}
    \begin{subfigure}{1.0\linewidth}
      \includegraphics[width=\linewidth]{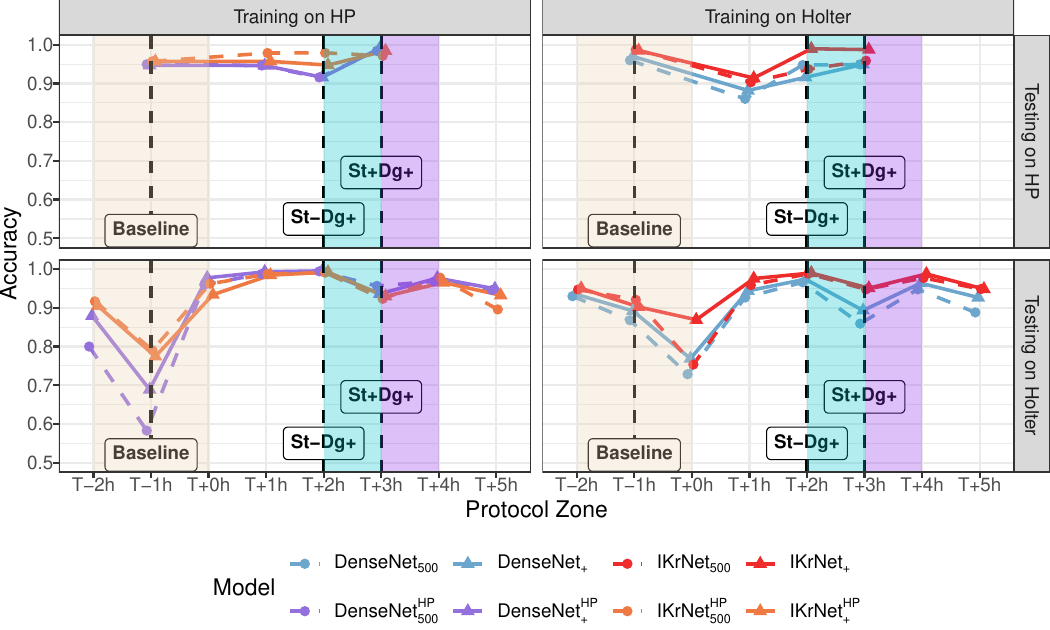}
      \subcaption{}
      \label{fig:accuracy_per_pt_holter}
    \end{subfigure}
  \end{minipage}
  \caption{\textbf{Cross-dataset evaluation of \texttt{IKrNet} and \texttt{DenseNet}} on different sampling rates (a), and across full protocol at 500 Hz (b)}
  \label{fig:hrv_analysis}
\end{figure*}

This section describes and evaluates eight models, summarized in~\suppref{tab:notation}. Specifically, we assess the performance of the state-of-the-art (SOTA) model, {\densenetbasic} for Sotalol footprint detection, as described in~\cite{prifti2021deep}, concerning its behavior on ECGs downsampled at different sampling frequencies and its robustness across the protocol zone analyzed in~\cref{sec:framework}. In their paper,~\cite{prifti2021deep} trained and tested on {\generepol}. Their model was trained for 100 epochs using the Adam optimizer, with a learning rate of 0.001 and a dropout rate of 0.2.
We considered three variations of the model based on the \texttt{DenseNet} architecture. These models are trained on {\generepol} with data augmentation ({\densenet}), {\generepolholter} alone ({\densenetbasicholter}), and {\generepolholter} with data augmentation ({\densenetholter}). Finally, we evaluated the proposed \texttt{IKrNet} architecture, considering models trained on {\generepol} (\ikrnetbasic), {\generepol} with data augmentation ({\ikrnet}), {\generepolholter} alone ({\ikrnetbasicholter}), and {\generepolholter} with data augmentation ({\ikrnetholter}).

\subsection{Performance analysis of models}
\paragraph{Analysis on {\generepol} and {\generepolholter}.}
The results in~\cref{fig:accuracy_per_sr_all_datasets} show that 
when tested as in the original setup on the {\generepol} dataset at 500 Hz, {\densenetbasic} performed very closely (within a one percentage point difference) to the results claimed in the original paper, achieving an accuracy of 95.77\%, a precision of 96.03\%, a recall of 96.41\%, and an F1 score of 96.22\%.

Those results also reveal a drop in performance when {\densenetbasic} is tested on the {\generepolholter} dataset, regardless of the sampling frequency ($\sim$2 percentage points difference between the accuracy of {\densenetbasic} on ECGs at 500 Hz from {\generepol} and the same setting in {\generepolholter}). However, {\densenetbasicholter} is more stable across multiple sampling rates and shows higher accuracy than {\densenetbasic} when tested on {\generepolholter}. From~\cref{fig:accuracy_per_sr_all_datasets}, we also observe that {\ikrnetbasic} and {\ikrnetbasicholter} display higher accuracies compared to previous models and have similar performance on multiple sampling rates. When trained and tested on the {\generepol} dataset, {\ikrnetbasic} achieved an accuracy of 96.66\%, surpassing {\densenetbasic}’s 95.77\%. A similar improvement was observed on the {\generepolholter} dataset, where {\ikrnetbasicholter} reached 94.37\% in accuracy (higher than the 89.98\% accuracy of {\densenetbasicholter}). In~\suppref{tab:short-recording-patient-500,tab:long-recording-patient-500}, we provide performance results on cross-dataset evaluations.

\paragraph{Analysis on data-augmented ECGs.}
\texttt{DenseNet} models do not show consistency when tested in settings with variability in sampling frequency (cf.~\cref{fig:accuracy_per_sr_all_datasets} and~\suppref{tab:short-recording-500,tab:long-recording-500}). Specifically, we observed a decrease in accuracy performance by up to approximately 9\% when resampling the signal from 500 Hz to 150 Hz then to 500 Hz. Similarly, the F1 score decreased up to approximately 7 percentage points. This performance drop could be expected due to the inevitable loss of information during downsampling. To assess whether the performance drop of {\densenetbasic} was due to the architecture itself or some epistemic uncertainty caused by the lack of data in the training set, we retrained the models by considering ECGs resampled at different sampling frequencies, as explained in~\cref{sec:dataset}. The numerical results for this second set of experiments are in \suppref{tab:short-recording-patient-+,tab:long-recording-patient-+}.

The experiments confirmed that augmenting the training dataset significantly improves the model's generalization capability across different sampling rates. On ECGs at 150 Hz, accuracy increased from 86.08\% with {\densenetbasic} to 90.31\% with {\densenet} when testing on {\generepol}. However, the results on {\generepolholter} remain unsatisfactory, with {\densenet} failing to achieve an overall accuracy greater than 70\% (cf.~\cref{fig:hrv_analysis}). Conversely, {\densenetbasicholter} and {\densenetholter}, benefiting from the diversity in the {\generepolholter} training set, achieved accuracies of 87.41\% and 89.58\% on {\generepolholter} at 150 Hz, respectively.
The accuracy of \texttt{IKrNet} improved by considering more sampling rates when training. For instance, {\ikrnetbasic} accuracy increased from 93.43\% to 95.10\% with {\ikrnet} when testing on {\generepol}. On the {\generepolholter} dataset, {\ikrnetbasicholter} accuracy improved from 90.30\% to 92.46\% at 150 Hz with {\ikrnetholter} model. At 500 Hz on the same dataset, {\ikrnetholter} has an accuracy of \textbf{94.00\%} outperforming {\densenetholter}’s 91.90\%, and when tested on the {\generepol} dataset, they achieved accuracies of \textbf{97.88\%} and 94.77\% respectively. \texttt{IKrNet} demonstrated greater performance compared to \texttt{DenseNet}, consistently outperforming it across different datasets and conditions.

\paragraph{Robustness to the protocol zones.}
\cref{tab:results_steps_all} shows the numerical results across the protocol zones (cf.~\cref{sec:framework}). The Accuracy Parity Difference (APD) notably revealed model instability across protocol zones. On {\generepol}, all models performed similarly, with an APD difference of less than 7\% between the worst-performing model ({\densenetbasic}) and the best ({\ikrnetholter}). In contrast, on {\generepolholter}, models, particularly those trained on {\generepol}, show a significant performance decline. {\densenetbasic} exhibits a nearly 60\% accuracy gap between the {\baseline} and {\nostdg} results. Interestingly, {\ikrnetholter} consistently outperformed the other models in this protocol time zone. This can be explained by several factors: \textit{(i)} Although the training set from {\generepolholter} contains more perturbations compared to {\generepol}, it helps the model to better generalize and reduces overfitting to the specific dataset. This is further supported by the fact that \textit{all} models trained on this dataset generally show better performance. \textit{(ii)} The data augmentation technique also seems to play a key role, particularly as the models were tested on various sampling frequencies that were not present during training. \textit{(iii)}
The multi-resolution CNN in \texttt{IKrNet} which can adapt to different waveform segments helps it in detecting baseline variations, including subtle QT prolongation changes. In contrast, the fixed kernel size of {\texttt{DenseNet}} limits its ability to focus on shorter or longer sections of the waveform.

\begin{figure}[h]
    \centering
    \includegraphics[width=\columnwidth]{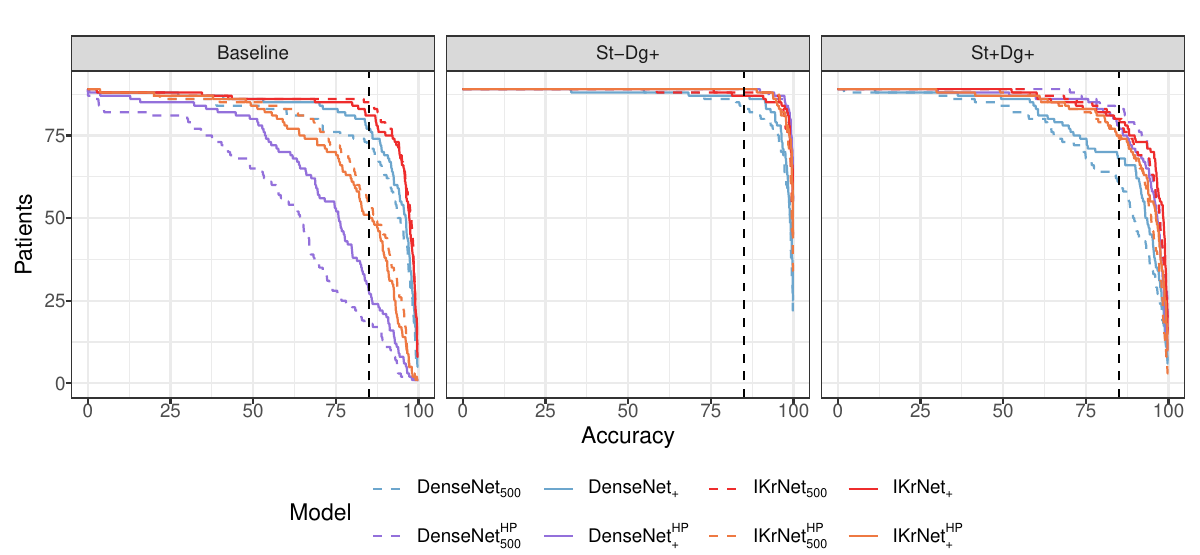}
    \caption{\textbf{Results on {\generepolholter} (ECGs at 500 Hz) over protocol zones and patients}. Number of patients (vertical axis) achieving accuracy above the threshold (horizontal axis). }
    \label{fig:auc_patients}
\end{figure}

In~\cref{fig:auc_patients}, to further investigate the model performances from a patient perspective, we considered all the ECGs recorded during the specified protocol periods for each model we measured the number of patients achieving accuracy above the examined threshold accuracy value (horizontal axis). 
The {\texttt{IKrNet}} models trained on {\generepolholter} exhibit consistent performance across study participants, with more than 75 out of 90 (83\%), in the holdout set, achieving an accuracy higher than 85\% (red lines). In contrast, the other models showed more variability. Notably, {\densenetbasic} performed poorly in the {\baseline} zone, with fewer than 25 out of 90 (28\%) participants with an accuracy above 85\% (violet dashed line).

{\densenetbasic} mistakes the most when considering the ECGs recorded one hour before Sotalol intake, as seen in~\cref{fig:patient_zoom_panel_b}. During this period, the model's accuracy was less than 85\% for fewer than 25 out of 90 participants (27\%).

Interestingly, some of the {\densenetbasic} limitations were also reported in~\citet{granese2023negative}, where it was demonstrated that the model lacked robustness when ECGs are processed through denoising methods, even after retraining.

\subsection{Heart rate effect on drug footprint inference}
\begin{figure*}[h!]
  \centering
  \begin{minipage}{0.48\linewidth}
    \begin{subfigure}{1.0\linewidth}
      \includegraphics[width=\linewidth]{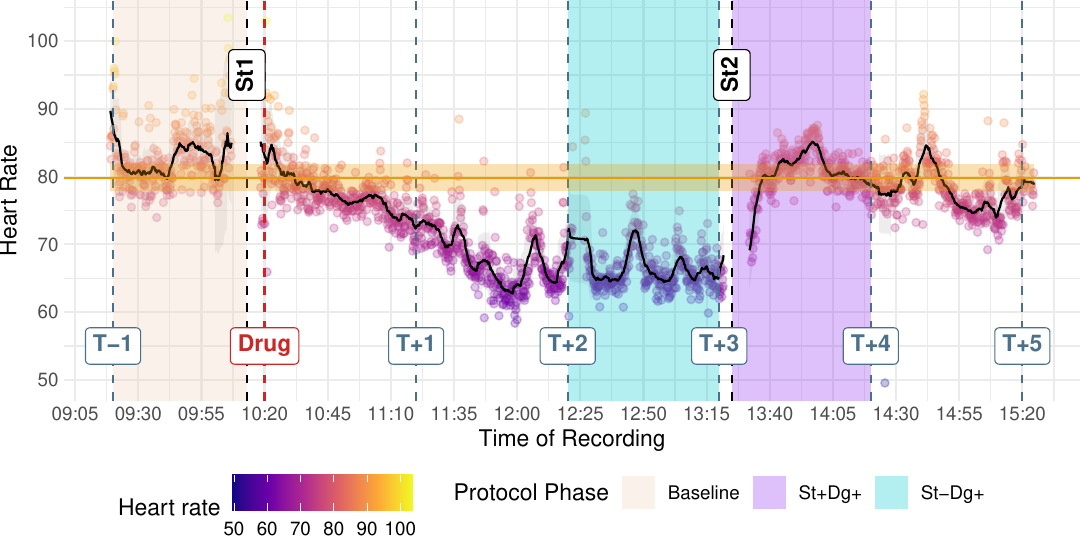}
      \subcaption{Heart rate variability.}
      \label{fig:patient_zoom_panel_a}
    \end{subfigure}
    
    \vspace{0.3cm}
    
    \begin{subfigure}{1.0\linewidth}
      \includegraphics[width=\linewidth]{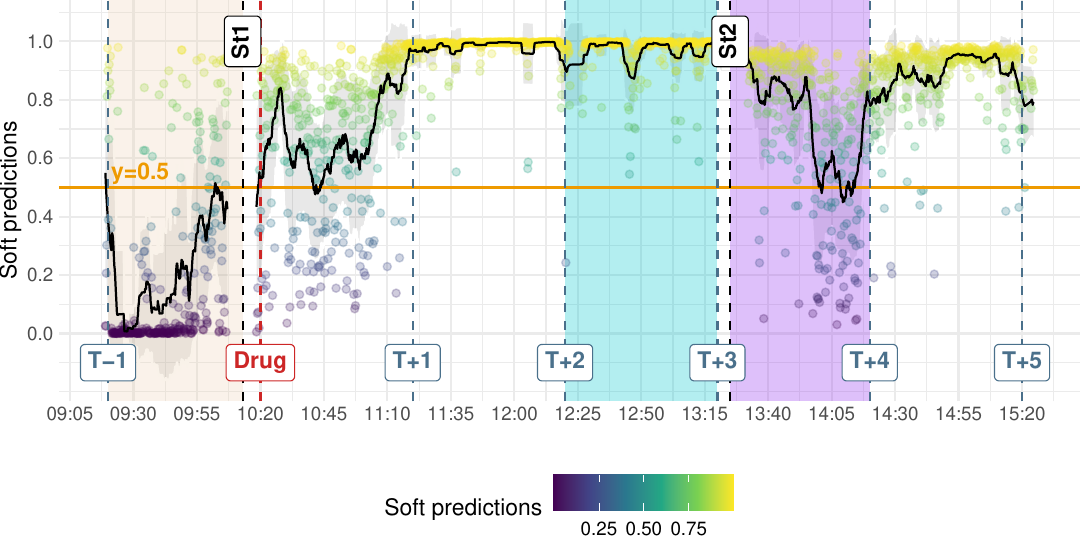}
      \subcaption{\densenetbasic predictions.}
      \label{fig:patient_zoom_panel_b}
    \end{subfigure}
    
    \vspace{0.3cm}
    
    \begin{subfigure}{1.0\linewidth}
      \includegraphics[width=\linewidth]{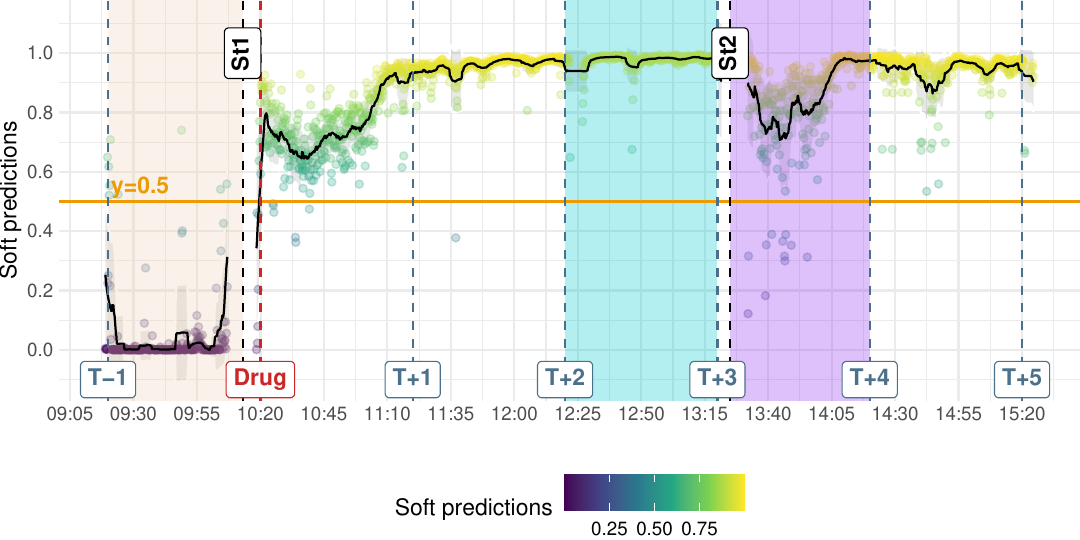}
      \subcaption{\ikrnetholter predictions.}
      \label{fig:patient_zoom_panel_c}
    \end{subfigure}
  \end{minipage}
  \hfill
  \begin{minipage}{0.48\linewidth}
    \begin{minipage}{0.48\linewidth}
        \begin{subfigure}{1.0\linewidth}
          \includegraphics[width=\linewidth]{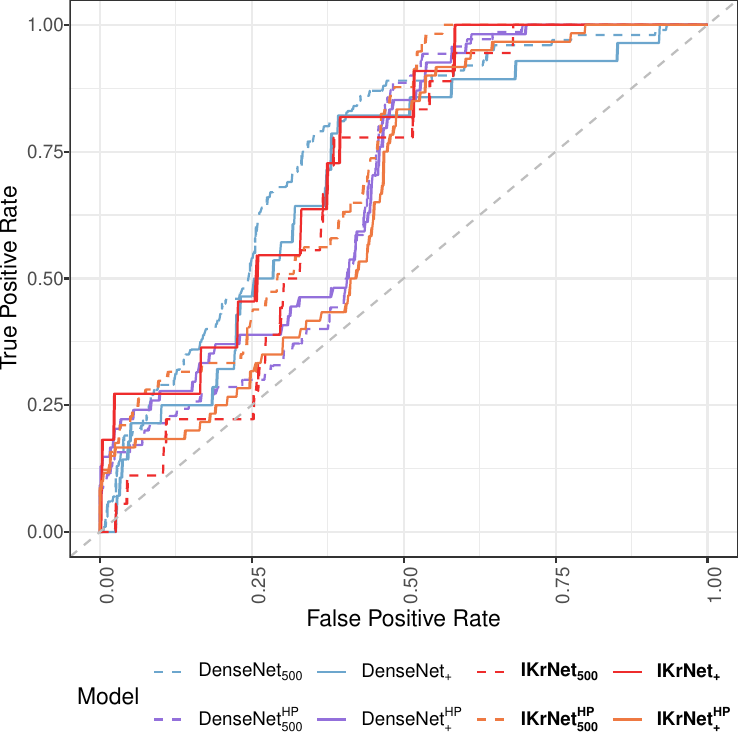}
          \subcaption{ROC curves with heart rate as a predictor of misclassifications}
          \label{fig:patient_zoom_panel_2}
        \end{subfigure}
    \end{minipage}
    \begin{minipage}{0.48\linewidth}
        \begin{subfigure}{1.0\linewidth}
          \includegraphics[width=\linewidth]{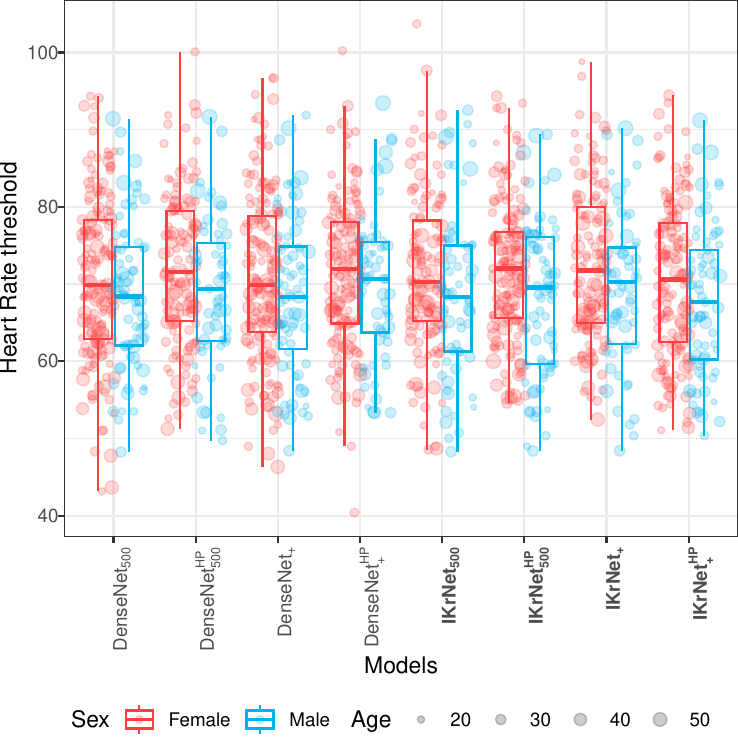}
          \subcaption{Heart Rate threshold per model on all patients}
          \label{fig:patient_zoom_panel_e}
        \end{subfigure}
    \end{minipage}
    
    \vspace{0.3cm}
    
    \begin{subfigure}{1.0\linewidth}
      \includegraphics[width=\linewidth]{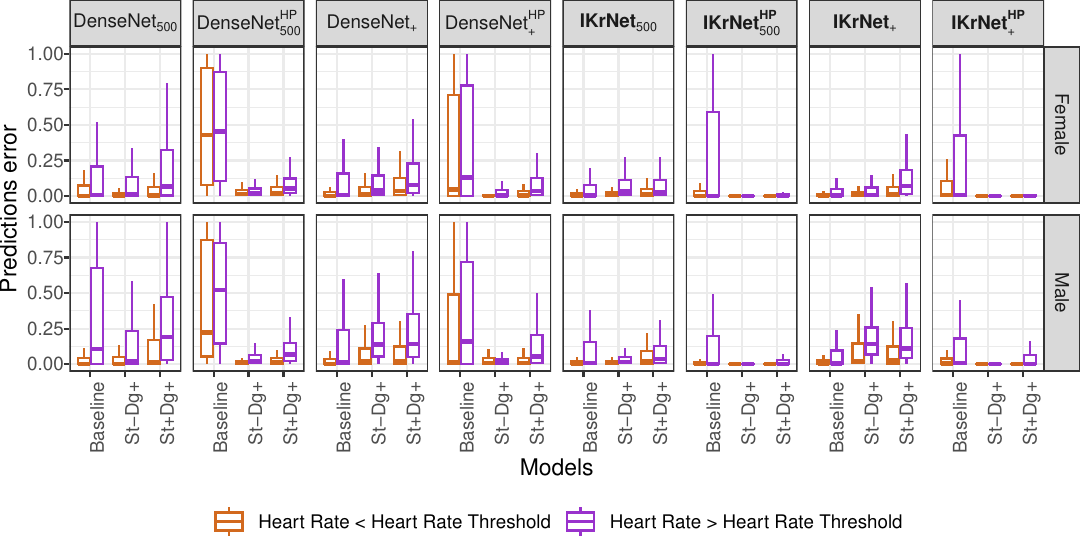}
      \subcaption{Model's prediction errors by heart rate groups on all sampling rates}
      \label{fig:patient_zoom_panel_f}
    \end{subfigure}
    
    \vspace{0.3cm}
    
    \begin{subfigure}{1.0\linewidth}
      \includegraphics[width=\linewidth]{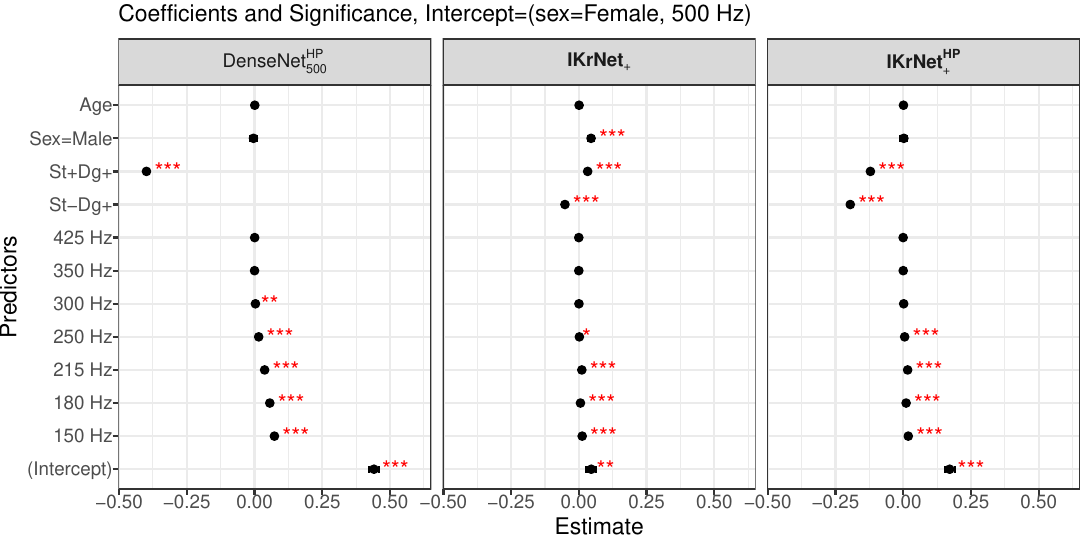}
      \subcaption{LME models~\cite{pinheiro2000linear}: $predictions\_error \sim\ sampling\_rate + sex + age + (1 | patient)$ }
      \label{fig:patient_zoom_panel_g}
    \end{subfigure}
  \end{minipage}
  \caption{\textbf{{\densenetbasic} and {\ikrnetholter} on single patient and influence of heart rate variability across time:} {\densenetbasic} is negatively affected by sampling rate changes and stress zones where heart rate increases, whereas {\ikrnetholter} is more robust.}
  \label{fig:hrv_analysis}
\end{figure*}

Heart rate variability (HRV)~\cite{Rajendra_Acharya2006-ye} is an important factor that can affect model performance, particularly in real-world scenarios where physiological states fluctuate over time~\cite{kotriwar2018higherorderspectralanalysis}. Sotalol is known to lower heart rate~\cite{HOHNLOSER1993A67} (cf.~\cref{fig:protocol}, zone {\nostdg}), an effect that models can typically detect. However, sudden increases in heart rate due to external factors among which stress efforts (cf.~\cref{fig:protocol}, zone {\stdg}), can obscure the drug's effect, potentially confusing the models and impacting their predictive accuracy. In~\cref{fig:hrv_analysis}, we illustrate a comparison  between \densenetbasic and \ikrnetholter, focusing on data from a single study participant across the entire protocol duration. We assessed how HRV impacted both models' performance and how effectively each model can adapt to varying heart rate conditions.

\paragraph{Identifying optimal heart rate (OHR) thresholds} We identified heart rate thresholds by analyzing the ROC curves for each model, as shown in~\cref{fig:patient_zoom_panel_2}, using heart rate as a predictor of the models' misclassification errors. By calculating the Youden Index~\cite{https://doi.org/10.1002/bimj.200410135} on the ROC curves, which maximize the difference between sensitivity and specificity, we determined the optimal heart rate threshold for each model and each study participant, beyond which models start to struggle. Interestingly, we observed a high variability of OHR throughout the study participants, as illustrated by boxplots in~\cref{fig:patient_zoom_panel_e}. This was partially explained by gender and age. 

\paragraph{Predictions errors analysis} Additionally, we computed the prediction errors for the ECGs above the OHR and below and observed that the predicted error was systematically higher above the OHR. Overall this error was shown to be lower for the \texttt{IKrNet} models compared with \texttt{DenseNet} ~\cref{fig:patient_zoom_panel_f}. Furthermore, we computed Linear Mixed Effects models (LME)~\cite{pinheiro2000linear} to further assess the impact of sampling rates and stress on the models. \texttt{IKrNet} models show relatively stable performance across low- and high-heart rate conditions in all zones, particularly at {\stdg} where drug influence and physiological stress are present. In contrast, \texttt{DenseNet} models exhibit increased variability, with more prediction errors at {\baseline} but also at higher heart rates during the {\nostdg} and {\stdg} zones.
\section{Conclusion}

In this work, we presented \texttt{IKrNet}, a novel neural network architecture, to detect drug-induced patterns in ECGs. This architecture is tailored explicitly for ECG analysis, leveraging a multi-resolution CNN backbone with varying receptive fields to capture low- and high-frequency features. By integrating spatial feature extraction through the CNN backbone with temporal dependency analysis via BiLSTM layers, \texttt{IKrNet} effectively adapts to the complex, time-evolving patterns, which are typical of ambulatory ECG Holter recordings. These are affected by varying heart rate and an integration of different footprints induced by external stressors.
We considered a clinical use case where study participants were administered Sotalol, a pharmacological agent used to induce QT prolongation and mimic arrhythmogenic conditions that can lead to Torsades-de-Pointes. 
Our contributions include establishing a formal framework for model robustness to evaluate consistency quantitatively under heart rate variability and stress conditions. This framework offers a standardized approach to assess model reliability across diverse and fluctuating physiological states, which is critical for real-world applications.

\section*{Acknowledgments}
This study was supported by the ANR-20-CE17-0022 DeepECG4U funding from the French National Research Agency (ANR).

{
    \small
    \bibliographystyle{ieeenat_fullname}
    \bibliography{main}
}

\appendix
\clearpage
\setcounter{page}{1}
\maketitlesupplementary

\section{Heart rate estimation}
\label{app:heart_rate_estimation}
The heart rate refers to the number of times the heart beats per minute (bpm). It can be estimated 
from the time difference between two successive heartbeats as follows:
\begin{multline}
  \hr(\x_i, \x_{i+1}) = \frac{60 F_s}{ n_{R_{i+1}} - n_{R_{i}}}\\\text{with,~} n_{R_i} = \argmax_{n \in [n_{i-1},n_i]} \left|\x_i[n]\right|
\end{multline}
$n_{R_i}$ being the sample index of the R impulse local peak position in the signal pattern $x_i$.
This value can be averaged over the whole signal made of $K$ observed heartbeats as:
\begin{equation}
  \hr_{\text{avg}}(\x) = \frac{1}{K-1} \sum_{i=1}^{K-1} \hr(\x_i, \x_{i+1}).
\end{equation}

\section{Additional material to~\cref{sec:experimental_setting}}
\subsection{Notation}
\Cref{tab:notation} provides an overview of the model names referenced throughout the paper.
\begin{table}[h!]
    \centering
    \caption{\textbf{Table of notations for the experiments}.}
    \begin{tabular}{l|p{.65\columnwidth}}
    \toprule
         \textbf{Symbol} & \textbf{Description} \\
    \midrule
     {\texttt{DenseNet}$^{\text{HP}}_{\{\cdot\}}$} & SOTA DenseNet trained on {\generepol} with ECGs at 500 Hz as in~\cite{prifti2021deep} ($\{\cdot\} = 500$), or [180, 215 and 500] Hz ($\{\cdot\} = +$).\\
     \hline
    {\texttt{DenseNet}$_{\{\cdot\}}$} & SOTA DenseNet trained on {\generepolholter} with ECGs at 500 Hz ($\{\cdot\} = 500$), or [180, 215 and 500] Hz ($\{\cdot\} = +$).\\
     \hline
    {\texttt{IKrNet}$^{\text{HP}}_{\{\cdot\}}$} & IKrNet trained on {\generepol} with ECGs at 500 Hz ($\{\cdot\} = 500$), or [180, 215 and 500] Hz ($\{\cdot\} = +$).\\
     \hline
    {\texttt{IKrNet}$_{\{\cdot\}}$} & IKrNet trained on {\generepolholter} with ECGs at 500 Hz ($\{\cdot\} = 500$), or [180, 215 and 500] Hz ($\{\cdot\} = +$).\\
    \bottomrule
    \end{tabular}
    \label{tab:notation}
\end{table}

\subsection{Datasets}
\Cref{tab:generepol_ecg_distribution} describes the dataset per partition. The number of samples in the table refers to the total amount after data augmentation as described in~\cref{sec:dataset}. 

\begin{table*}[h!]
  \centering
  \caption{\textbf{{\generepolholter} and {\generepol} datasets}. 
The amount of raw signal (at 500 Hz) is obtained by dividing each partition value by 3, except in Holdout, where it is divided by 8. $F_s$ represents the sampling rates to augment the datasets.}
  \begin{tabular}{l|l|p{2cm}||c|c|c|c}
    \toprule
     \multicolumn{3}{c||}{\textbf{Partition details}} & 
     \multicolumn{2}{c|}{\textbf{\generepol}} & \multicolumn{2}{c}{\textbf{\generepolholter}}\\
    \midrule
     \multirow{2}{*}{\textbf{Name}} & \multirow{2}{*}{\textbf{N°Patients}} & \multirow{2}{*}{\textbf{$F_s$ (Hz)}} & \textbf{Sot-} &\textbf{Sot+} & \textbf{Sot-} &\textbf{Sot+}\\
     & & & \multicolumn{2}{c|}{\textit{Raw / Augmented}} & \multicolumn{2}{c|}{\textit{Raw / Augmented}} \\
    \midrule
    Training & 629 & 180, 250, 500 & 2739/8217 & 3576/10728 & 30736/92207 & 30808/92424\\
    \midrule
    Validation & 90 &180, 250, 500 & 413/1239 & 539/1617 & 4698/14092 & 4739/14216 \\
    \midrule
    Evaluation & 90 & 180, 250, 500 & 339/1017 & 451/1353 & 4898/14692 & 3287/9860\\
    \midrule
    Holdout & 90 & 150, 180, 215, 250, 300, 350, 425, 500 & 396/3168 & 502/4016 & 4673/37384 & 4710/37680 \\
    \bottomrule
  \end{tabular}
  \label{tab:generepol_ecg_distribution}
\end{table*}


\subsection{IKrNet architecture}
\begin{figure}[h]
    \centering
    \includegraphics[width=0.9\columnwidth]{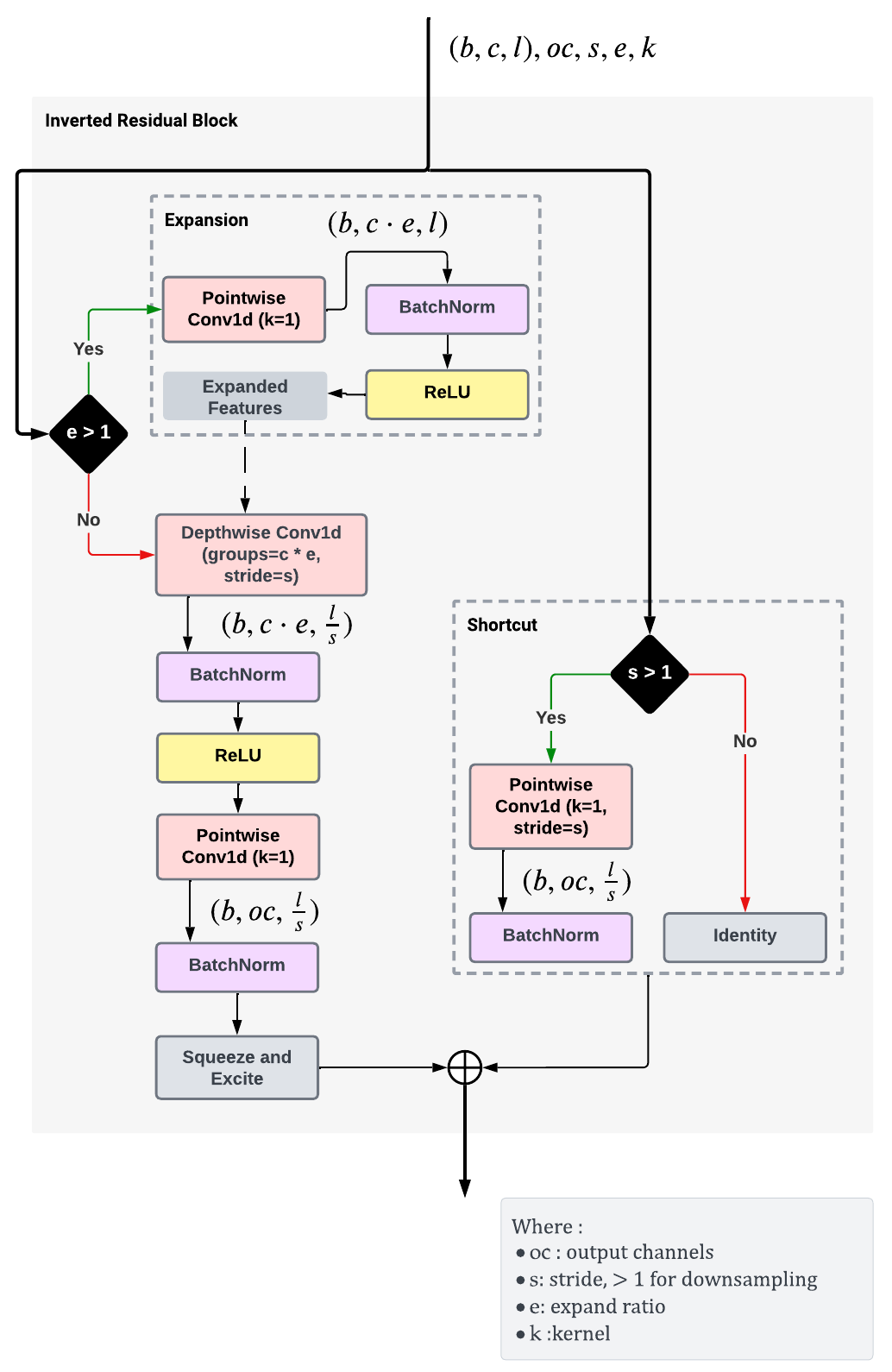}
    \caption{\textbf{Inverted residual block} structure enables optimal feature extraction while reducing the complexity of traditional basic blocks.}
    \label{fig:model_architecture_inverted_residual_block}
\end{figure}

\begin{figure}[h]
    \centering
    \includegraphics[width=.8\columnwidth]{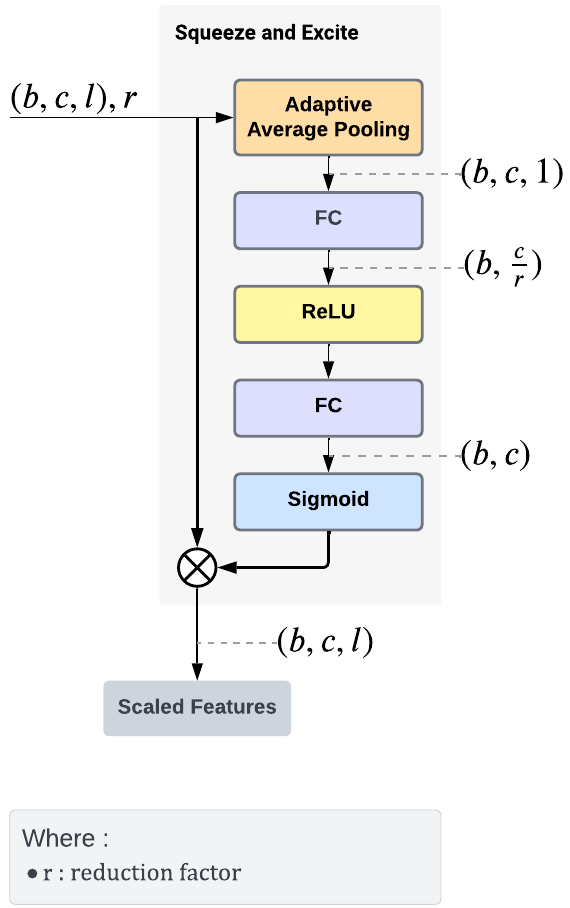}
    \caption{\textbf{Squeeze-and-Excite Block.} This structure is designed to enhance model performance by dynamically re-calibrating feature channels and optimizing feature extraction and prioritization within the network.}
    \label{fig:model_architecture_inverted_squeeze_and_excite}
\end{figure}

\Cref{fig:model_architecture_inverted_residual_block,fig:model_architecture_inverted_squeeze_and_excite} provide a detailed view of the Inverted Residual Block and Squeeze-and-Excite Block modules. We refer to~\cref{sec:ikrnet} for the general architecture description.

\section{Additional material to~\cref{sec:results}}

\subsection{LME models to evaluate models per zone}
\begin{figure*}[h]
    \centering 
    \includegraphics[width=0.9\linewidth]{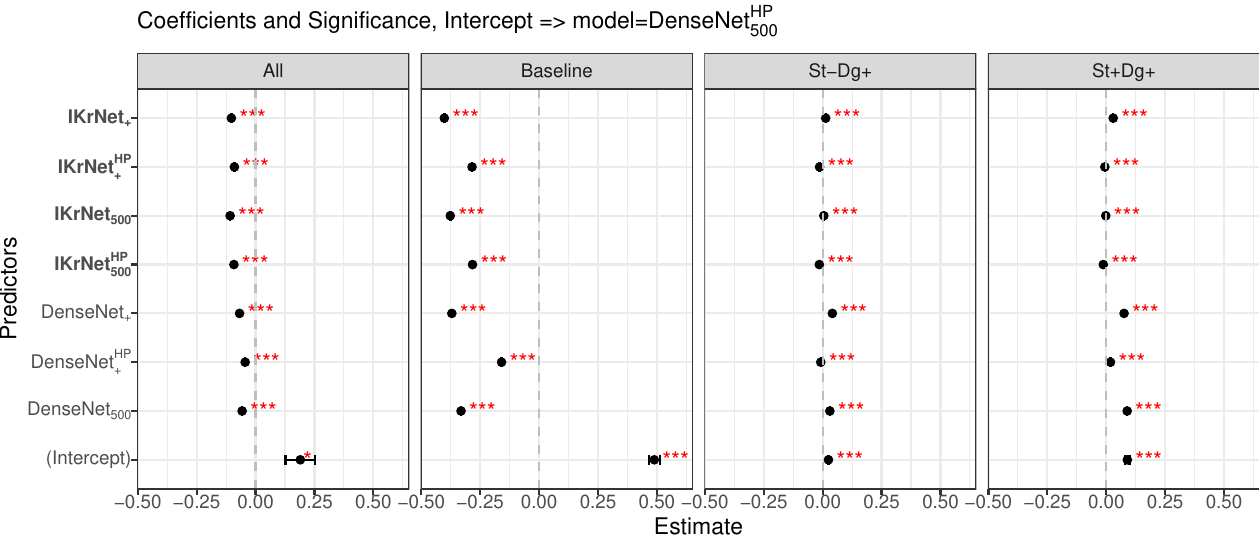} 
    \caption{Linear Mixed Effects models per zone on all sampling rates to evaluate model predictions errors. In \textit{all zones} plot (first facet) we used model: $predictions\_error ~ model + (1 | sampling\_rate) + (1 | zone) + (1 | patient\_id) + (1 | ecg\_id)$. For each zone individually, we used the model: $predictions\_error ~ model + (1 | sampling\_rate) + (1 | patient\_id) + (1 | ecg\_id)$.}
    \label{fig:lme_models_per_zone} 
\end{figure*}

In~\cref{fig:lme_models_per_zone} we build 4 Linear Mixed Effects models to evaluate the prediction errors of all models across protocol zones and on all sampling rates. Results confirm observations from~\cref{tab:results_steps_all} where \texttt{DenseNet} models perform poorly on the {\baseline} zone compared to {\texttt{IKrNet}} models. {\ikrnetholter} is generally more robust when considering all zones.

\subsection{Ablation study}
We conducted an ablation study in which we evaluated different configurations of the \texttt{IKrNet} architecture in the protocol zones. Specifically, we evaluated the number of resolution branches in the CNN backbone, the type of convolution block in each branch (Inverted Residual Block~\cref{fig:model_architecture_inverted_residual_block}, or basic block), the output size of each branch, the number of BiLSTM layers and the use of the residual skip connection leading to the combination of spatial and temporal features.

Each configuration is trained on the {\generepolholter} dataset with data augmentation. Results of the best-performing configurations are presented in~\Cref{tab:ablation_study_results}; the first row corresponds to the final architecture structure that yields the best performances. 

We observed the model's performance drop when reducing the number of resolution branches in the CNN backbone to capture spatial features. This suggests that a less diverse set of receptive fields diminishes the model's ability to detect abrupt signal changes, as it limits the range of spatial patterns the model can effectively analyze. Moreover, we observed that increasing the output size of each branch reduces the accuracy and leads to overfitting. BiLSTM layers also play a critical role; when removed from the architecture, we notice a significant drop in accuracy, particularly in the Sot + class in the {\nostdg} and {\stdg} zones. This highlights the importance of capturing temporal dependencies, as BiLSTM layers enable the model to analyze the progression of ECG signals over time. Without these layers, the architecture struggles between transient heart rate induced changes and sustained drug-induced patterns, leading to a higher misclassification rate.
The skip connection used to combine spatial and temporal features is also crucial for maintaining robust model performance, we noticed a drop of the accuracy by 7\% in the {\stdg} zone, where stress coincides with the drug's effects. The skip connection ensures that critical spatial features, such as subtle morphological changes in the ECG waveform, are retained and integrated with temporal patterns captured by the BiLSTM layers. Without this connection, the model struggles to effectively link the localized spatial features with their temporal progression.

\begin{table*}[t]
  \centering
  \caption{\textbf{Ablation study results:} different configurations are evaluated on the protocol zones. Each configuration is trained on {\generepolholter} with data-augmentation}
  \begin{tabular}{c|c|c|c|c||c|c|c}
    \toprule
    \multicolumn{5}{c||}{\textbf{Configurations}} & \multicolumn{3}{c}{\textbf{Accuracy}} \\
    \midrule
    \vtop{\hbox{\strut N° resolution}\hbox{\strut branches}} & \vtop{\hbox{\strut Type of conv}\hbox{\strut block}} & \vtop{\hbox{\strut Output size per}\hbox{\strut resolution branch}} & \vtop{\hbox{\strut N° BiLSTM}\hbox{\strut layers}} & \vtop{\hbox{\strut Use residual}\hbox{\strut skip link}} & \baseline & \nostdg & \stdg \\
    \midrule
    \textbf{4} & \textbf{\vtop{\hbox{\strut Inverted}\hbox{\strut residual}}} & \textbf{256} & \textbf{2} & \cmark & \textbf{92.22} & \textbf{98.92} & \textbf{95.35} \\
    \midrule
    4 & \vtop{\hbox{\strut Inverted}\hbox{\strut residual}} & 256 & 2 & \xmark & 92.45 & 91.01 & 88.23 \\
    \midrule
    4 & \vtop{\hbox{\strut Inverted}\hbox{\strut residual}} & 256 & 1 & \cmark & 91.92 & 96.34 & 92.67 \\
    \midrule
    4 & \vtop{\hbox{\strut Inverted}\hbox{\strut residual}} & 256 & 0 & \cmark & 97.89 & 85.90 & 82.04 \\
    \midrule
    2 & \vtop{\hbox{\strut Inverted}\hbox{\strut residual}} & 256 & 2 & \cmark & 90.65 & 92.93 & 88.67 \\
    \midrule
    1 & \vtop{\hbox{\strut Inverted}\hbox{\strut residual}} & 256 & 2 & \cmark & 89.31 & 89.94 & 85.34 \\
    \midrule
    4 & \vtop{\hbox{\strut Inverted}\hbox{\strut residual}} & 512 & 2 & \cmark & 92.06 & 96.68 & 94.19 \\
    \midrule
    4 & Basic & 256 & 2 & \cmark & 90.25 & 94.12 & 92.57 \\
    \bottomrule
  \end{tabular}
  \label{tab:ablation_study_results}
\end{table*}
    
\label{sec:ablation_study}

\subsection{Per sampling-rates results}
\begin{figure*}[h!]
  \centering
  \begin{minipage}{0.48\linewidth}
    \begin{subfigure}{1.0\linewidth}
      \includegraphics[width=\linewidth]{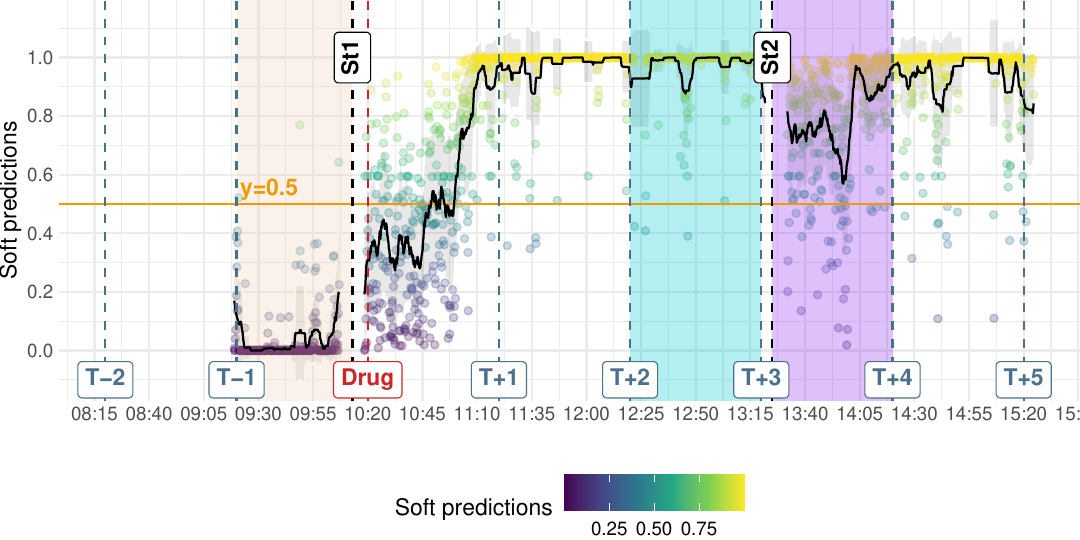}
      \subcaption{{\densenetbasicholter} predictions.}
      \label{fig:patient_zoom_models_panel_a}
    \end{subfigure}
    
    \vspace{0.3cm}
    
    \begin{subfigure}{1.0\linewidth}
      \includegraphics[width=\linewidth]{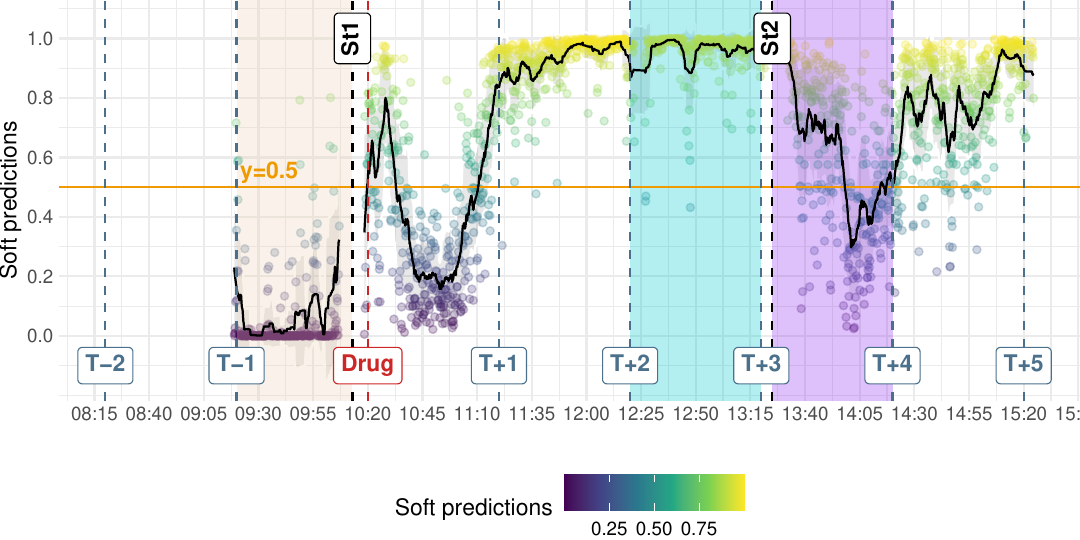}
      \subcaption{{\densenetholter} predictions.}
      \label{fig:patient_zoom_models_panel_b}
    \end{subfigure}
    
    \vspace{0.3cm}
    
    \begin{subfigure}{1.0\linewidth}
      \includegraphics[width=\linewidth]{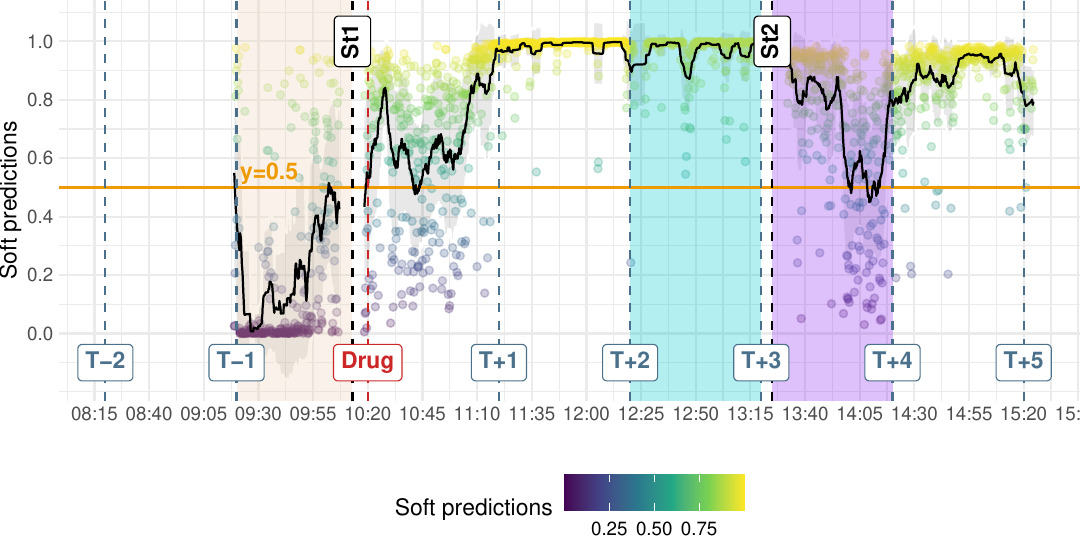}
      \subcaption{{\densenetbasic} predictions.}
      \label{fig:patient_zoom_models_panel_c}
    \end{subfigure}

    \vspace{0.3cm}
    
    \begin{subfigure}{1.0\linewidth}
      \includegraphics[width=\linewidth]{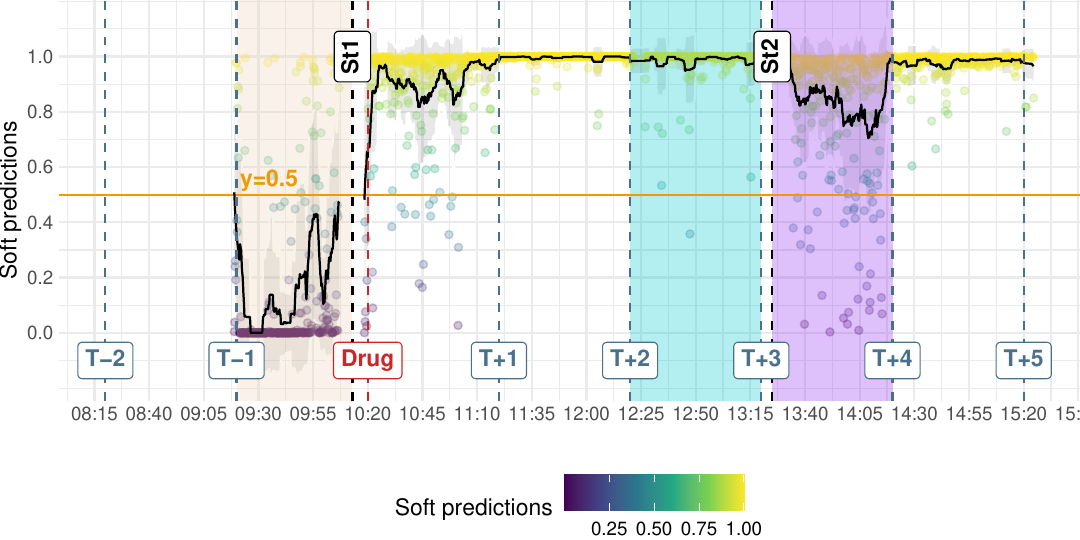}
      \subcaption{{\densenet} predictions.}
      \label{fig:patient_zoom_models_panel_d}
    \end{subfigure}
  \end{minipage}
  \hfill
  \begin{minipage}{0.48\linewidth}
    \begin{subfigure}{1.0\linewidth}
      \includegraphics[width=\linewidth]{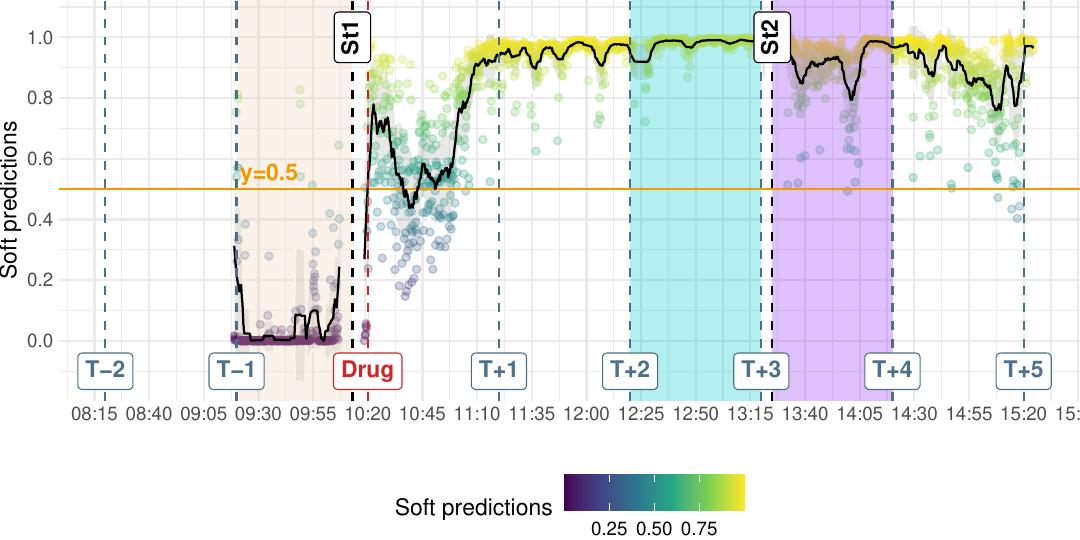}
      \subcaption{{\ikrnetbasicholter} predictions.}
      \label{fig:patient_zoom_models_panel_e}
    \end{subfigure}
    
    \vspace{0.3cm}
    
    \begin{subfigure}{1.0\linewidth}
      \includegraphics[width=\linewidth]{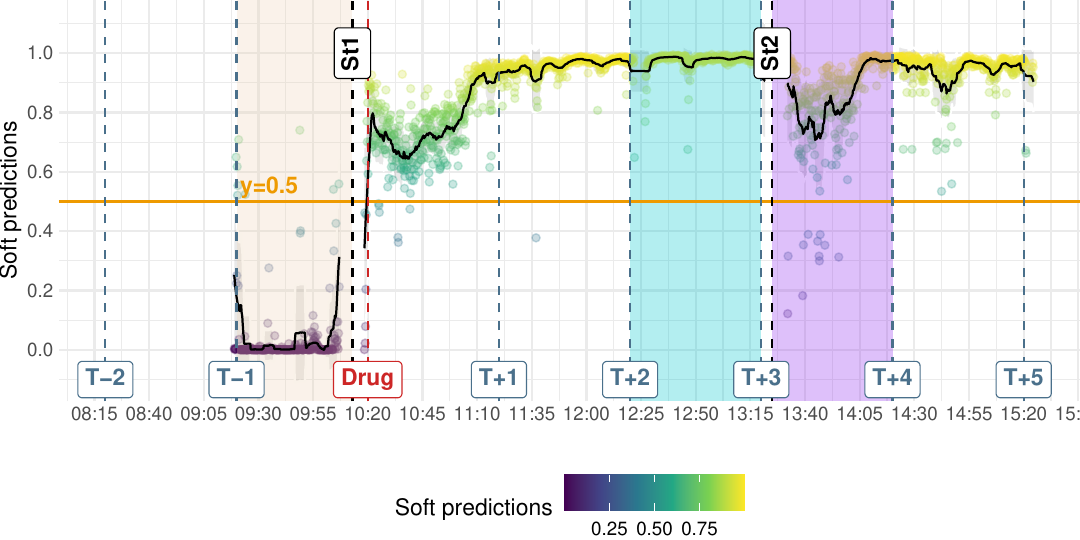}
      \subcaption{{\ikrnetholter} predictions.}
      \label{fig:patient_zoom_models_panel_f}
    \end{subfigure}
    
    \vspace{0.3cm}
    
    \begin{subfigure}{1.0\linewidth}
      \includegraphics[width=\linewidth]{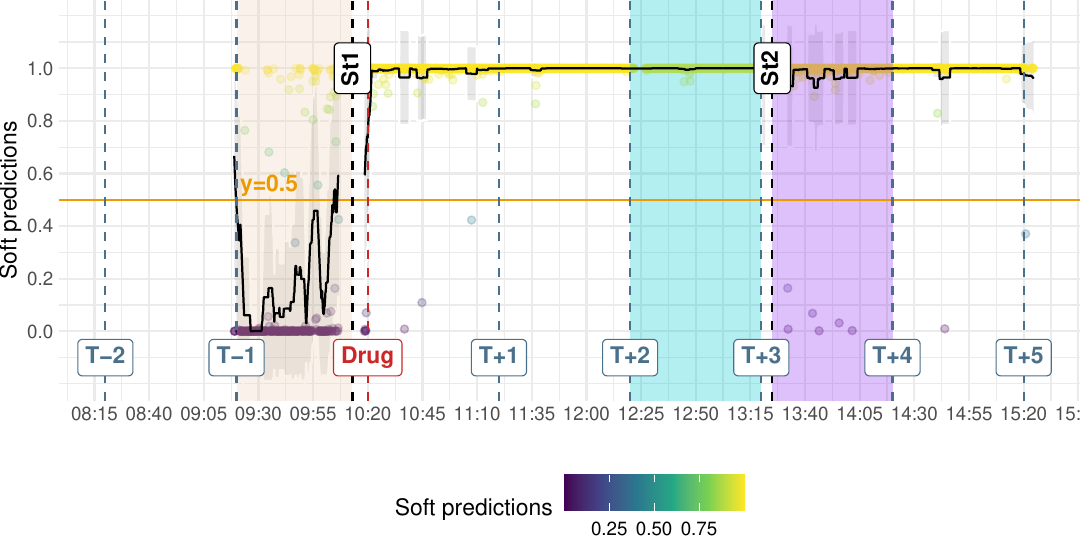}
      \subcaption{{\ikrnetbasic} predictions.}
      \label{fig:patient_zoom_models_panel_g}
    \end{subfigure}

    \vspace{0.3cm}
    
    \begin{subfigure}{1.0\linewidth}
      \includegraphics[width=\linewidth]{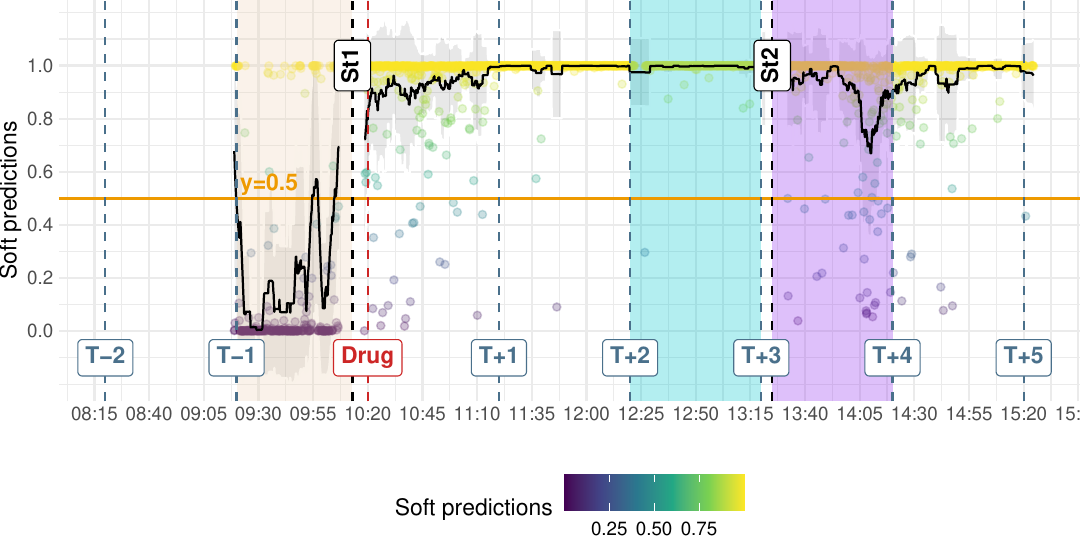}
      \subcaption{{\ikrnet} predictions.}
      \label{fig:patient_zoom_models_panel_h}
    \end{subfigure}
  \end{minipage}
  \caption{All models predictions on a single patient.}
  \label{fig:all_models_on_single_patient}
\end{figure*}

\Cref{tab:short-recording-500,tab:short-recording-+} present the numerical results of the eight models trained on the {\generepol} dataset; \Cref{tab:long-recording-500,tab:long-recording-+} summarize the results of the eight models trained on the {\generepolholter} datasets. In~\cref{fig:all_models_on_single_patient} we tested all eight models on a single patient similar to~\cref{fig:hrv_analysis}. Finally,~\cref{tab:short-recording-patient-500,tab:short-recording-patient-+,tab:long-recording-patient-500,tab:long-recording-patient-+} present the results per patient and averaged over all the patients.

In this set of experiments, we evaluated models' performance under varying ECG sampling rates without distinguishing results by protocol zones as in~\cref{sec:results}. Overall, the \texttt{IKrNet} architecture consistently outperforms the \texttt{DenseNet} architecture, regardless of the training or testing set.

Similar to the robustness setting of the protocol zones, we assessed the models' robustness by defining subgroups based on ECGs at different sampling rates:
$acc_{\text{func}}\myeq\text{func}\{acc(\texttt{150}), acc(\texttt{180}),
acc(\texttt{215}),
acc(\texttt{250}),
acc(\texttt{300}),$ $acc(\texttt{350}),
acc(\texttt{425}),
acc(\texttt{500})\}$, with $\text{func}\in\{\min, \max\}$. Below, we indicate the accuracy parity difference (APD) results for this setting. Results indicate that IKrNet+, trained in both datasets, is the most robust model among all frequency sampling rates.

\paragraph{APD$\downarrow_{0}$ of the models tested on {\generepol}}
\begin{itemize}
    \item {\densenetbasic}: 9.69 (Max accuracy: 95.99)
    \item {\densenet}: 5.57 (Max accuracy: 95.99)
    \item {\densenetbasicholter}: 2.33 (Max accuracy: 94.65)
    \item {\densenetholter}: \underline{\textbf{0.67} (Max accuracy: 94.88)}
    \item {\ikrnetbasic}: 3.45 (Max accuracy: 96.88)
    \item {\ikrnet}: 1.45 (Max accuracy: 96.55)
    \item {\ikrnetbasicholter}: 1.89 (Max accuracy: 96.99)
    \item {\ikrnetholter}: \textbf{0.9} \textbf{(Max accuracy: 98)}
\end{itemize}

\paragraph{APD$\downarrow_{0}$ of the models tested on {\generepolholter}}
\begin{itemize}
    \item {\densenetbasic}: 16.51 (Max accuracy: 79.47)
    \item {\densenet}: 17.84 (Max accuracy: 84.80)
    \item {\densenetbasicholter}: 2.57 (Max accuracy: 89.98)
    \item {\densenetholter}: 2.34 (Max accuracy: 91.92)
    \item {\ikrnetbasic}: 7.14 (Max accuracy: 88.51)
    \item {\ikrnet}: 2.27 (Max accuracy: 88.19)
    \item {\ikrnetbasicholter}: \underline{4.07 \textbf{(Max accuracy: 94.37)}}
    \item {\ikrnetholter}: \textbf{1.54} \textbf{(Max accuracy: 94)}
\end{itemize}

\begin{table*}[h!]
    \centering
    \caption{\textbf{Comparison between {\densenetbasic}
    and {\ikrnetbasic}}. The models are trained on {\generepol} with ECGs at 500 Hz. The best accuracy is highlighted in \textbf{bold}, and the best F1 score is highlighted with \underline{underline}.}
    \begin{tabular}{c|c|c|c|c||c|c|c|c}
    \toprule
       \multirow{3}{*}{\textbf{$F_s$}} & \multicolumn{4}{c||}{\textbf{\densenetbasic}} & \multicolumn{4}{c}{\textbf{\ikrnetbasic}} \\
        \cmidrule{2-9}
        & Accuracy & Precision & Recall & F1 Score & Accuracy & Precision & Recall & F1 Score \\
          \midrule
          \multicolumn{9}{c}{\textbf{Test on {\generepol} dataset}} \\
          \midrule
         150 & 86.08 & 80.75 & 98.61 & 88.79 & \textbf{93.43} & 90.49 & 98.61 & \underline{94.38} \\
         180 & 90.98 & 86.74 & 99.00 & 92.47 & \textbf{94.88} & 93.35 & 97.81 & \underline{95.53} \\
         215 & \textbf{93.99} & 91.33 & 98.61 & \underline{94.83} & 93.10 & 90.74 & 97.61 & 94.05 \\
         250 & 94.88 & 94.36 & 96.61 & 95.47 & \textbf{96.55} & 96.27 & 97.61 & \underline{96.93} \\
         300 & 95.32 & 95.63 & 96.02 & 95.83 & \textbf{96.55} & 96.27 & 97.61 & \underline{96.93} \\
         350 & \textbf{95.99} & 96.05 & 96.81 & 96.43 & 96.88 & 96.65 & 97.81 & \underline{97.23} \\
         425 & 95.43 & 95.64 & 96.22 & 95.93 & \textbf{96.88} & 96.65 & 97.81 & \underline{97.23} \\
         500 & 95.77 & 96.03 & 96.41 & 96.22 & \textbf{96.66} & 96.64 & 97.41 & \underline{97.02} \\
          \midrule
          \multicolumn{9}{c}{\textbf{Test on {\generepolholter} dataset}} \\
          \midrule
         150 & 62.96 & 57.58 & 99.23 & 72.88 & \textbf{81.37} & 73.90 & 97.16 & \underline{83.95} \\
         180 & 66.53 & 60.06 & 99.28 & 74.84 & \textbf{85.24} & 78.70 & 96.75 & \underline{86.80} \\
         215 & 70.68 & 63.28 & 98.88 & 77.17 & \textbf{84.10} & 77.35 & 96.57 & \underline{85.90} \\
         250 & 75.95 & 68.01 & 98.26 & 80.38 & \textbf{87.85} & 82.42 & 96.30 & \underline{88.82} \\
         300 & 78.80 & 71.04 & 97.42 & 82.17 & \textbf{88.26} & 83.07 & 96.19 & \underline{89.15} \\
         350 & 79.47 & 71.74 & 97.42 & 82.63 & \textbf{88.34} & 83.16 & 96.21 & \underline{89.21} \\
         425 & 79.28 & 71.50 & 97.58 & 82.53 & \textbf{88.38} & 83.22 & 96.21 & \underline{89.25} \\
         500 & 79.66 & 71.98 & 97.31 & 82.75 & \textbf{88.51} & 83.54 & 95.99 & \underline{89.34} \\
    \bottomrule    
    \end{tabular}
    \label{tab:short-recording-500}
\end{table*}

\begin{table*}[h!]
    \centering
    \caption{\textbf{Comparison between {\densenet}
    and {\ikrnet}}. The models are trained on {\generepol} augmented with ECGs at 180, 215, and 500 Hz. The best accuracy is highlighted in \textbf{bold}, and the best F1 score is highlighted with \underline{underline}.}
    \begin{tabular}{c|c|c|c|c||c|c|c|c}
    \toprule
       \multirow{3}{*}{\textbf{$F_s$}} & \multicolumn{4}{c||}{\textbf{\densenet}} & \multicolumn{4}{c}{\textbf{\ikrnet}} \\
        \cmidrule{2-9}
        & Accuracy & Precision & Recall & F1 Score & Accuracy & Precision & Recall & F1 Score \\
          \midrule
          \multicolumn{9}{c}{\textbf{Test on {\generepol} dataset}} \\
          \midrule
         150 & 90.31 & 85.71 & 99.20 & 91.97 & \textbf{95.10} & 95.80 & 95.42 & \underline{95.61} \\
         180 & 95.10 & 93.87 & 97.61 & 95.70 & \textbf{95.77} & 95.85 & 96.61 & \underline{96.23} \\
         215 & 94.77 & 93.17 & 97.81 & 95.43 & \textbf{95.10} & 94.55 & 96.81 & \underline{95.67} \\
         250 & 94.99 & 94.89 & 96.22 & 95.55 & \textbf{95.99} & 95.87 & 97.01 & \underline{96.44} \\
         300 & 94.77 & 95.23 & 95.42 & 95.32 & \textbf{96.21} & 96.43 & 96.81 & \underline{96.62} \\
         350 & 95.88 & 95.68 & 97.01 & 96.34 & \textbf{96.55} & 96.45 & 97.41 & \underline{96.93} \\
         425 & 95.99 & 95.87 & 97.01 & 96.44 & \textbf{96.55} & 96.63 & 97.21 & \underline{96.92} \\
         500 & 95.88 & 95.86 & 96.81 & 96.33 & \textbf{96.55} & 96.63 & 97.21 & \underline{96.92} \\
          \midrule
          \multicolumn{9}{c}{\textbf{Test on {\generepolholter} dataset}} \\
          \midrule
         150 & 66.96 & 60.40 & 99.05 & 75.04 & \textbf{85.97} & 80.36 & 95.32 & \underline{87.20} \\
         180 & 78.92 & 71.05 & 97.82 & 82.31 & \textbf{86.88} & 81.28 & 95.92 & \underline{88.00} \\
         215 & 80.77 & 73.13 & 97.44 & 83.56 & \textbf{85.92} & 79.91 & 96.08 & \underline{87.25} \\
         250 & 83.62 & 76.52 & 97.15 & 85.60 & \textbf{87.47} & 82.02 & 96.08 & \underline{88.50} \\
         300 & 84.41 & 77.59 & 96.90 & 86.17 & \textbf{87.91} & 82.58 & 96.18 & \underline{88.86} \\
         350 & 84.80 & 77.97 & 97.11 & 86.50 & \textbf{88.12} & 82.79 & 96.35 & \underline{89.05} \\
         425 & 84.73 & 77.84 & 97.21 & 86.46 & \textbf{88.19} & 82.85 & 96.41 & \underline{89.12} \\
         500 & 84.77 & 77.87 & 97.26 & 86.49 & \textbf{88.13} & 82.77 & 96.38 & \underline{89.06} \\
    \bottomrule    
    \end{tabular}
    \label{tab:short-recording-+}
\end{table*}

\begin{table*}[h!]
    \centering
    \caption{\textbf{Comparison between {\densenetbasicholter}
    and {\ikrnetbasicholter}}. The models are trained on {\generepol}  with ECGs at 500 Hz. The best accuracy is highlighted in \textbf{bold}, and the best F1 score is highlighted with \underline{underline}.}
    \begin{tabular}{c|c|c|c|c||c|c|c|c}
    \toprule
       \multirow{3}{*}{\textbf{$F_s$}} & \multicolumn{4}{c||}{\textbf{\densenetbasicholter}} & \multicolumn{4}{c}{\textbf{\ikrnetbasicholter}} \\
        \cmidrule{2-9}
          & Accuracy & Precision & Recall & F1 Score & Accuracy & Precision & Recall & F1 Score \\
          \midrule
          \multicolumn{9}{c}{\textbf{{Test on \generepol} dataset}} \\
          \midrule
          150 & 92.32 & 94.82 & 91.24 & 92.99 & \textbf{95.10} & 95.80 & 95.42 & \underline{95.61} \\
          180 & 92.76 & 94.14 & 92.83 & 93.48 & \textbf{96.33} & 98.15 & 95.22 & \underline{96.66} \\
          215 & 93.54 & 94.40 & 94.02 & 94.21 & \textbf{95.43} & 96.38 & 95.42 & \underline{95.90} \\
          250 & 94.32 & 95.93 & 93.82 & 94.86 & \textbf{96.99} & 98.77 & 95.82 & \underline{97.27} \\
          300 & 94.21 & 96.30 & 93.23 & 94.74 & \textbf{96.21} & 98.35 & 94.82 & \underline{96.55} \\
          350 & 94.54 & 96.70 & 93.43 & 95.04 & \textbf{96.88} & 98.97 & 95.42 & \underline{97.16} \\
          425 & 94.65 & 96.90 & 93.43 & 95.13 & \textbf{96.77} & 98.96 & 95.22 & \underline{97.06} \\
          500 & 94.43 & 96.69 & 93.23 & 94.93 & \textbf{96.33} & 98.96 & 94.42 & \underline{96.64} \\

          \midrule
          \multicolumn{9}{c}{\textbf{Test on {\generepolholter} dataset}} \\
          \midrule
          150 & 87.41 & 85.49 & 90.21 & 87.79 & \textbf{90.30} & 85.41 & 97.27 & \underline{90.96} \\
          180 & 87.69 & 85.05 & 91.54 & 88.18 & \textbf{92.17} & 88.42 & 97.10 & \underline{92.56} \\
          215 & 88.18 & 85.43 & 92.15 & 88.66 & \textbf{90.74} & 86.02 & 97.34 & \underline{91.33} \\
          250 & 88.93 & 86.71 & 92.01 & 89.28 & \textbf{93.51} & 90.65 & 97.07 & \underline{93.75} \\
          300 & 89.51 & 87.71 & 91.97 & 89.79 & \textbf{93.76} & 91.04 & 97.12 & \underline{93.98} \\
          350 & 89.84 & 88.17 & 92.10 & 90.09 & \textbf{93.87} & 91.19 & 97.16 & \underline{94.08} \\
          425 & 89.94 & 88.36 & 92.06 & 90.17 & \textbf{94.10} & 91.71 & 97.00 & \underline{94.28} \\
          500 & 89.98 & 88.68 & 91.73 & 90.18 & \textbf{94.37} & 92.69 & 96.38 & \underline{94.50} \\
    \bottomrule    
    \end{tabular}
    \label{tab:long-recording-500}
\end{table*}
\begin{table*}[ht]
    \centering
    \caption{\textbf{Comparison between {\densenetholter}
    and {\ikrnetholter}}. The models are trained on {\generepol} augmented with ECGs at 180, 215, and 500 Hz. The best accuracy is highlighted in \textbf{bold}, and the best F1 score is highlighted with \underline{underline}.}
    \begin{tabular}{c|c|c|c|c||c|c|c|c}
    \toprule
       \multirow{3}{*}{\textbf{$F_s$}} & \multicolumn{4}{c||}{\textbf{\densenetholter}} & \multicolumn{4}{c}{\textbf{\ikrnetholter}} \\
        \cmidrule{2-9}
          & Accuracy & Precision & Recall & F1 Score & Accuracy & Precision & Recall & F1 Score \\
          \midrule
          \multicolumn{9}{c}{\textbf{{Test on \generepol} dataset}} \\
          \midrule
          
             150 & 94.21 & 96.30 & 93.23 & 94.74 & \textbf{97.10} & 97.98 & 96.81 & \underline{97.39} \\
             180 & 94.54 & 96.13 & 94.02 & 95.07 & \textbf{97.66} & 98.20 & 97.61 & \underline{97.90} \\
             215 & 94.88 & 96.72 & 94.02 & 95.35 & \textbf{96.44} & 95.90 & 97.81 & \underline{96.84} \\
             250 & 94.32 & 96.88 & 92.83 & 94.81 & \textbf{98.00} & 98.59 & 97.81 & \underline{98.20} \\
             300 & 94.88 & 97.50 & 93.23 & 95.32 & \textbf{97.77} & 98.59 & 97.41 & \underline{98.00} \\
             350 & 94.77 & 97.30 & 93.23 & 95.22 & \textbf{97.88} & 98.79 & 97.41 & \underline{98.09} \\
             425 & 94.77 & 97.30 & 93.23 & 95.22 & \textbf{97.88} & 98.79 & 97.41 & \underline{98.09} \\
             500 & 94.77 & 97.49 & 93.03 & 95.21 & \textbf{97.88} & 98.79 & 97.41 & \underline{98.09} \\
          \midrule
          \multicolumn{9}{c}{\textbf{Test on {\generepolholter} dataset}} \\
          \midrule
          150 & 89.58 & 87.19 & 92.87 & 89.94 & \textbf{92.46} & 89.13 & 96.77 & \underline{92.79} \\
 180 & 90.40 & 87.59 & 94.21 & 90.78 & \textbf{93.03} & 89.68 & 97.29 & \underline{93.33} \\
 215 & 90.51 & 87.99 & 93.88 & 90.84 & \textbf{92.55} & 88.73 & 97.53 & \underline{92.92} \\
 250 & 91.28 & 89.40 & 93.71 & 91.50 & \textbf{93.69} & 90.80 & 97.27 & \underline{93.93} \\
 300 & 91.81 & 90.19 & 93.87 & 91.99 & \textbf{93.9}1 & 91.23 & 97.20 & \underline{94.12} \\
 350 & 91.77 & 90.18 & 93.79 & 91.95 & \textbf{93.98} & 91.31 & 97.24 & \underline{94.18} \\
 425 & 91.92 & 90.28 & 94.00 & 92.10 & \textbf{93.97} & 91.29 & 97.24 & \underline{94.17} \\
 500 & 91.90 & 90.37 & 93.84 & 92.07 & \textbf{94.00} & 91.42 & 97.14 & \underline{94.19} \\
    \bottomrule    
    \end{tabular}
    \label{tab:long-recording-+}
\end{table*}

\begin{table*}[h]
    \centering
    \caption{\textbf{Per patient comparison between {\densenetbasic} and {\ikrnetbasic}}. The models are trained on {\generepol} with ECGs at 500 Hz. The best accuracy is highlighted in \textbf{bold}, and the best F1 score is highlighted with \underline{underline}. Borderline cases are omitted, i.e., when the models predict only negative samples for a patient or when the patient contains only true negative samples.}
    \begin{tabular}{c|c|c|c|c||c|c|c|c}
    \toprule
        \multirow{3}{*}{\textbf{$F_s$}} & \multicolumn{4}{c||}{\textbf{\densenetbasic}} & \multicolumn{4}{c}{\textbf{\ikrnetbasic}} \\
         \cmidrule{2-9}
        & Accuracy  & Precision & Recall & F1 Score & Accuracy & Precision & Recall & F1 Score \\
        \midrule
        \multicolumn{9}{c}{\textbf{Test on {\generepol} dataset}} \\
        \midrule
 150 & 86.2 $\pm$ 20.9 & 81.5 $\pm$ 26.4 & 98.8 $\pm$ 4.9 & 91.5 $\pm$ 10.5 & \textbf{93.9} $\pm$ 13.5 & 91.7 $\pm$ 17.3 & 98.8 $\pm$ 4.5 & \underline{95.6} $\pm$ 7.8 \\
 180 & 91.4 $\pm$ 14.6 & 86.9 $\pm$ 23.3 & 99.1 $\pm$ 4.5 & 94.3 $\pm$ 9.0 & \textbf{94.8} $\pm$ 13.1 & 93.7 $\pm$ 16.4 & 98.0 $\pm$ 6.5 & \underline{96.4} $\pm$ 7.2 \\
 215 & \textbf{94.4} $\pm$ 12.7 & 92.1 $\pm$ 18.4 & 98.8 $\pm$ 5.1 & \underline{95.5} $\pm$ 9.6 & 92.4 $\pm$ 17.9 & 90.8 $\pm$ 21.9 & 97.9 $\pm$ 6.5 & 95.3 $\pm$ 9.5 \\
 250 & 95.5 $\pm$ 10.1 & 95.3 $\pm$ 13.4 & 97.0 $\pm$ 8.5 & 96.1 $\pm$ 7.5 & \textbf{96.5} $\pm$ 11.7 & 96.9 $\pm$ 12.4 & 97.9 $\pm$ 6.5 & \underline{97.5} $\pm$ 6.3 \\
 300 & 95.8 $\pm$ 10.4 & 96.3 $\pm$ 12.9 & 96.4 $\pm$ 10.0 & 96.3 $\pm$ 8.4 & \textbf{96.5} $\pm$ 11.8 & 96.9 $\pm$ 12.4 & 97.8 $\pm$ 6.7 & \underline{97.5} $\pm$ 6.3 \\
 350 & 96.3 $\pm$ 9.7 & 96.5 $\pm$ 12.8 & 97.2 $\pm$ 8.2 & 96.9 $\pm$ 7.3 & \textbf{96.8} $\pm$ 11.7 & 97.2 $\pm$ 12.3 & 98.1 $\pm$ 6.3 & \underline{97.8} $\pm$ 6.2 \\
 425 & 95.9 $\pm$ 10.4 & 96.2 $\pm$ 13.0 & 96.6 $\pm$ 9.5 & 96.4 $\pm$ 8.4 & \textbf{96.8} $\pm$ 11.7 & 97.2 $\pm$ 12.3 & 98.1 $\pm$ 6.4 & \underline{97.8} $\pm$ 6.2 \\
 500 & 96.1 $\pm$ 10.2 & 96.6 $\pm$ 12.8 & 96.8 $\pm$ 9.3 & 96.7 $\pm$ 8.1 & \textbf{96.6} $\pm$ 11.7 & 97.2 $\pm$ 12.3 & 97.7 $\pm$ 6.8 & \underline{97.6} $\pm$ 6.3 \\
          \midrule
          \multicolumn{9}{c}{\textbf{Test on {\generepolholter} dataset}} \\
          \midrule
 150 & 63.8 $\pm$ 12.4 & 59.6 $\pm$ 10.8 & 99.2 $\pm$ 1.7 & 73.9 $\pm$ 8.3 & \textbf{81.6} $\pm$ 9.0 & 75.6 $\pm$ 10.4 & 97.1 $\pm$ 3.3 & \underline{84.6} $\pm$ 6.8 \\
 180 & 67.3 $\pm$ 12.6 & 62.3 $\pm$ 11.4 & 99.3 $\pm$ 1.4 & 75.9 $\pm$ 8.5 & \textbf{85.4} $\pm$ 8.4 & 80.2 $\pm$ 9.7 & 96.8 $\pm$ 4.3 & \underline{87.3} $\pm$ 6.3 \\
 215 & 71.3 $\pm$ 12.3 & 65.5 $\pm$ 11.8 & 98.8 $\pm$ 2.1 & 78.2 $\pm$ 8.5 & \textbf{84.1} $\pm$ 9.5 & 79.1 $\pm$ 10.8 & 96.5 $\pm$ 4.3 & \underline{86.5} $\pm$ 6.8 \\
 250 & 76.3 $\pm$ 11.3 & 70.1 $\pm$ 11.6 & 98.2 $\pm$ 2.8 & 81.2 $\pm$ 7.9 & \textbf{87.9} $\pm$ 8.1 & 83.9 $\pm$ 9.7 & 96.3 $\pm$ 4.8 & \underline{89.2} $\pm$ 6.3 \\
 300 & 79.0 $\pm$ 11.1 & 73.1 $\pm$ 11.7 & 97.4 $\pm$ 3.6 & 82.9 $\pm$ 7.8 & \textbf{88.3} $\pm$ 7.9 & 84.5 $\pm$ 9.4 & 96.1 $\pm$ 4.9 & \underline{89.5} $\pm$ 6.1 \\
 350 & 79.7 $\pm$ 11.2 & 73.8 $\pm$ 11.8 & 97.4 $\pm$ 3.7 & 83.4 $\pm$ 7.9 & \textbf{88.4} $\pm$ 7.9 & 84.6 $\pm$ 9.5 & 96.2 $\pm$ 4.7 & \underline{89.6} $\pm$ 6.1 \\
 425 & 79.5 $\pm$ 10.9 & 73.6 $\pm$ 11.6 & 97.5 $\pm$ 3.5 & 83.3 $\pm$ 7.8 & \textbf{88.4} $\pm$ 7.9 & 84.6 $\pm$ 9.5 & 96.2 $\pm$ 4.8 & \underline{89.6} $\pm$ 6.1 \\
 500 & 79.9 $\pm$ 10.8 & 74.0 $\pm$ 11.5 & 97.3 $\pm$ 3.9 & 83.5 $\pm$ 7.7 & \textbf{88.5} $\pm$ 7.9 & 84.9 $\pm$ 9.4 & 95.9 $\pm$ 4.9 & \underline{89.7} $\pm$ 6.1 \\
    \bottomrule    
    \end{tabular}
    \label{tab:short-recording-patient-500}
\end{table*}
\begin{table*}[h!]
    \centering
    \caption{\textbf{Per patient comparison between {\densenet} and {\ikrnet}}. The models are trained on {\generepol} augmented with ECGs at 180, 215 and 500 Hz. The best accuracy is highlighted in \textbf{bold}, and the best F1 score is highlighted with \underline{underline}. Borderline cases are omitted, i.e., when the models predict only negative samples for a patient or when the patient contains only true negative samples.}
    \begin{tabular}{c|c|c|c|c||c|c|c|c}
    \toprule
        \multirow{3}{*}{\textbf{$F_s$}} & \multicolumn{4}{c||}{\textbf{\densenet}} & \multicolumn{4}{c}{\textbf{\ikrnet}} \\
         \cmidrule{2-9}
        & Accuracy  & Precision & Recall & F1 Score & Accuracy & Precision & Recall & F1 Score \\
        \midrule
        \multicolumn{9}{c}{\textbf{Test on {\generepol} dataset}} \\
          \midrule
 150 & 90.8 $\pm$ 17.0 & 87.7 $\pm$ 20.4 & 99.3 $\pm$ 3.7 & 93.8 $\pm$ 8.8 & \textbf{95.6} $\pm$ 9.1 & 97.2 $\pm$ 8.4 & 95.7 $\pm$ 9.6 & \underline{95.9} $\pm$ 7.2 \\
 180 & 95.3 $\pm$ 11.3 & 94.2 $\pm$ 16.6 & 97.8 $\pm$ 6.5 & 96.5 $\pm$ 7.5 & \textbf{96.3} $\pm$ 8.4 & 97.2 $\pm$ 8.4 & 97.0 $\pm$ 8.1 & \underline{96.6} $\pm$ 6.6 \\
 215 & 94.8 $\pm$ 13.3 & 93.5 $\pm$ 17.5 & 98.1 $\pm$ 5.7 & \underline{96.1} $\pm$ 8.5 & \textbf{95.0} $\pm$ 11.2 & 94.4 $\pm$ 16.9 & 97.1 $\pm$ 8.1 & 96.0 $\pm$ 8.3 \\
 250 & 95.1 $\pm$ 12.5 & 95.8 $\pm$ 13.1 & 95.7 $\pm$ 12.2 & 96.3 $\pm$ 7.0 & \textbf{96.5} $\pm$ 8.5 & 97.3 $\pm$ 8.3 & 97.3 $\pm$ 7.7 & \underline{96.9} $\pm$ 6.4 \\
 300 & 95.1 $\pm$ 11.2 & 96.0 $\pm$ 13.1 & 95.6 $\pm$ 10.7 & 95.7 $\pm$ 8.8 & \textbf{96.6} $\pm$ 8.6 & 97.7 $\pm$ 8.1 & 97.0 $\pm$ 8.2 & \underline{96.9} $\pm$ 6.6 \\
 350 & 95.9 $\pm$ 12.4 & 96.5 $\pm$ 12.8 & 97.3 $\pm$ 7.0 & 97.0 $\pm$ 6.7 & \textbf{96.6} $\pm$ 9.4 & 96.9 $\pm$ 12.3 & 97.7 $\pm$ 7.6 & \underline{97.3} $\pm$ 6.4 \\
 425 & 96.0 $\pm$ 12.4 & 96.6 $\pm$ 12.8 & 97.3 $\pm$ 7.0 & 97.1 $\pm$ 6.7 & \textbf{96.6} $\pm$ 9.5 & 97.0 $\pm$ 12.3 & 97.5 $\pm$ 7.8 & \underline{97.3} $\pm$ 6.4 \\
 500 & 95.9 $\pm$ 12.4 & 96.6 $\pm$ 12.8 & 97.1 $\pm$ 7.3 & 97.0 $\pm$ 6.8 & \textbf{96.6} $\pm$ 9.5 & 97.0 $\pm$ 12.3 & 97.5 $\pm$ 7.8 & \underline{97.3} $\pm$ 6.4 \\
          \midrule
          \multicolumn{9}{c}{\textbf{Test on {\generepolholter} dataset}} \\
          \midrule
 150 & 67.9 $\pm$ 12.1 & 62.6 $\pm$ 10.8 & 99.0 $\pm$ 2.2 & 76.1 $\pm$ 8.3 & \textbf{86.1} $\pm$ 8.2 & 82.0 $\pm$ 10.2 & 95.2 $\pm$ 6.6 & \underline{87.6} $\pm$ 6.7 \\
 180 & 79.3 $\pm$ 10.3 & 73.0 $\pm$ 11.1 & 97.8 $\pm$ 3.4 & 83.1 $\pm$ 7.5 & \textbf{87.0} $\pm$ 8.0 & 82.9 $\pm$ 10.0 & 95.8 $\pm$ 6.2 & \underline{88.4} $\pm$ 6.5 \\
 215 & 80.9 $\pm$ 10.1 & 74.9 $\pm$ 11.0 & 97.5 $\pm$ 4.0 & 84.2 $\pm$ 7.3 & \textbf{86.0} $\pm$ 8.7 & 81.6 $\pm$ 10.3 & 96.0 $\pm$ 6.3 & \underline{87.6} $\pm$ 6.8 \\
 250 & 83.8 $\pm$ 9.2 & 78.1 $\pm$ 10.4 & 97.2 $\pm$ 4.2 & 86.1 $\pm$ 6.9 & \textbf{87.6} $\pm$ 8.0 & 83.5 $\pm$ 9.9 & 96.0 $\pm$ 6.4 & \underline{88.8} $\pm$ 6.5 \\
 300 & 84.6 $\pm$ 9.2 & 79.2 $\pm$ 10.4 & 96.9 $\pm$ 4.0 & 86.7 $\pm$ 6.9 & \textbf{88.0} $\pm$ 7.8 & 84.1 $\pm$ 9.7 & 96.1 $\pm$ 6.2 & \underline{89.2} $\pm$ 6.4 \\
 350 & 84.9 $\pm$ 9.3 & 79.5 $\pm$ 10.4 & 97.1 $\pm$ 4.0 & 87.0 $\pm$ 6.9 & \textbf{88.2} $\pm$ 7.9 & 84.3 $\pm$ 9.7 & 96.3 $\pm$ 6.0 & \underline{89.4} $\pm$ 6.4 \\
 425 & 84.8 $\pm$ 9.3 & 79.4 $\pm$ 10.4 & 97.2 $\pm$ 3.9 & 87.0 $\pm$ 6.9 & \textbf{88.3} $\pm$ 7.8 & 84.3 $\pm$ 9.7 & 96.3 $\pm$ 5.9 & \underline{89.4} $\pm$ 6.4 \\
 500 & 84.9 $\pm$ 9.3 & 79.4 $\pm$ 10.3 & 97.3 $\pm$ 3.7 & 87.0 $\pm$ 6.9 & \textbf{88.2} $\pm$ 7.9 & 84.3 $\pm$ 9.8 & 96.3 $\pm$ 5.9 & \underline{89.4} $\pm$ 6.4 \\
    \bottomrule    
    \end{tabular}
    \label{tab:short-recording-patient-+}
\end{table*}

\begin{table*}[h!]
    \centering
    \caption{\textbf{Per patient comparison between {\densenetbasicholter} and {\ikrnetbasicholter}}. The models are trained on {\generepolholter} with ECGs at 500 Hz. The best accuracy is highlighted in \textbf{bold}, and the best F1 score is highlighted with \underline{underline}. Borderline cases are omitted, i.e., when the models predict only negative samples for a patient or when the patient contains only true negative samples.}
    \begin{tabular}{c|c|c|c|c||c|c|c|c}
    \toprule
       \multirow{3}{*}{\textbf{$F_s$}} & \multicolumn{4}{c||}{\textbf{\densenetbasicholter}} & \multicolumn{4}{c}{\textbf{\ikrnetbasicholter}} \\
        \cmidrule{2-9}
          & Accuracy  & Precision & Recall & F1 Score & Accuracy & Precision & Recall & F1 Score \\
          \midrule
          \multicolumn{9}{c}{\textbf{Test on {\generepol} dataset}} \\
          \midrule
150 & 92.8 $\pm$ 13.4 & 95.8 $\pm$ 13.5 & 91.5 $\pm$ 16.9 & 92.7 $\pm$ 12.9 & \textbf{95.4} $\pm$ 11.1 & 96.2 $\pm$ 12.4 & 95.9 $\pm$ 13.1 & \underline{95.7} $\pm$ 10.4 \\
180 & 93.2 $\pm$ 13.5 & 95.3 $\pm$ 13.5 & 93.3 $\pm$ 15.8 & 93.4 $\pm$ 12.5 & \textbf{96.8} $\pm$ 8.1 & 98.6 $\pm$ 5.3 & 95.8 $\pm$ 12.0 & \underline{96.6} $\pm$ 8.4 \\
215 & 93.4 $\pm$ 14.7 & 94.0 $\pm$ 18.5 & 94.7 $\pm$ 13.0 & 94.9 $\pm$ 10.1 & \textbf{95.3} $\pm$ 13.1 & 96.2 $\pm$ 15.0 & 96.1 $\pm$ 11.8 & \underline{96.4} $\pm$ 8.8 \\
250 & 94.5 $\pm$ 11.5 & 95.7 $\pm$ 15.4 & 94.3 $\pm$ 13.1 & 95.1 $\pm$ 9.7 & \textbf{97.0} $\pm$ 8.9 & 98.2 $\pm$ 10.5 & 96.4 $\pm$ 11.4 & \underline{97.3} $\pm$ 8.1 \\
300 & 94.4 $\pm$ 12.6 & 96.8 $\pm$ 12.2 & 93.7 $\pm$ 14.8 & 94.7 $\pm$ 11.2 & \textbf{96.5} $\pm$ 9.7 & 97.9 $\pm$ 10.8 & 95.5 $\pm$ 13.6 & \underline{96.4} $\pm$ 10.4 \\
350 & 94.8 $\pm$ 12.1 & 96.9 $\pm$ 12.6 & 94.0 $\pm$ 13.0 & 95.2 $\pm$ 10.0 & \textbf{97.4} $\pm$ 8.1 & 99.2 $\pm$ 4.3 & 96.0 $\pm$ 12.8 & \underline{96.9} $\pm$ 9.6 \\
425 & 94.9 $\pm$ 12.1 & 97.1 $\pm$ 12.5 & 94.0 $\pm$ 13.0 & 95.3 $\pm$ 10.0 & \textbf{97.3} $\pm$ 8.4 & 99.1 $\pm$ 4.6 & 95.8 $\pm$ 12.9 & \underline{96.8} $\pm$ 9.8 \\
500 & 94.6 $\pm$ 12.2 & 96.9 $\pm$ 12.6 & 93.7 $\pm$ 13.5 & 95.0 $\pm$ 10.4 & \textbf{96.9} $\pm$ 8.9 & 99.1 $\pm$ 4.6 & 95.1 $\pm$ 14.2 & \underline{96.3} $\pm$ 10.6 \\
          \midrule
          \multicolumn{9}{c}{\textbf{Test on {\generepolholter} dataset}} \\
          \midrule
150 & 87.6 $\pm$ 9.4 & 88.2 $\pm$ 12.0 & 90.0 $\pm$ 13.5 & 87.7 $\pm$ 10.7 & \textbf{90.5} $\pm$ 8.0 & 87.1 $\pm$ 10.4 & 97.2 $\pm$ 5.3 & \underline{91.4} $\pm$ 6.6 \\
180 & 87.9 $\pm$ 9.2 & 87.8 $\pm$ 12.1 & 91.4 $\pm$ 12.1 & 88.3 $\pm$ 9.6 & \textbf{92.3} $\pm$ 7.3 & 89.8 $\pm$ 9.3 & 97.0 $\pm$ 5.4 & \underline{92.9} $\pm$ 6.2 \\
215 & 88.4 $\pm$ 9.2 & 88.1 $\pm$ 12.0 & 92.0 $\pm$ 11.7 & 88.8 $\pm$ 9.3 & \textbf{90.7} $\pm$ 8.9 & 87.8 $\pm$ 10.9 & 97.3 $\pm$ 5.2 & \underline{91.8} $\pm$ 6.9 \\
250 & 89.1 $\pm$ 8.9 & 89.2 $\pm$ 11.6 & 91.8 $\pm$ 11.6 & 89.4 $\pm$ 9.0 & \textbf{93.6} $\pm$ 7.1 & 91.9 $\pm$ 8.8 & 97.0 $\pm$ 5.4 & \underline{94.0} $\pm$ 6.0 \\
300 & 89.7 $\pm$ 8.6 & 90.0 $\pm$ 11.2 & 91.8 $\pm$ 11.5 & 89.8 $\pm$ 8.9 & \textbf{93.9} $\pm$ 7.0 & 92.3 $\pm$ 8.7 & 97.0 $\pm$ 5.4 & \underline{94.2} $\pm$ 5.9 \\
350 & 90.0 $\pm$ 8.5 & 90.4 $\pm$ 10.9 & 92.0 $\pm$ 11.5 & 90.1 $\pm$ 8.8 & \textbf{94.0} $\pm$ 7.0 & 92.4 $\pm$ 8.8 & 97.1 $\pm$ 5.3 & \underline{94.3} $\pm$ 6.0 \\
425 & 90.1 $\pm$ 8.4 & 90.6 $\pm$ 10.9 & 91.9 $\pm$ 11.4 & 90.2 $\pm$ 8.7 & \textbf{94.2} $\pm$ 7.0 & 92.9 $\pm$ 8.6 & 96.9 $\pm$ 5.6 & \underline{94.5} $\pm$ 6.0 \\
500 & 90.1 $\pm$ 8.4 & 90.9 $\pm$ 10.9 & 91.6 $\pm$ 11.5 & 90.2 $\pm$ 8.7 & \textbf{94.5} $\pm$ 6.9 & 93.9 $\pm$ 8.4 & 96.3 $\pm$ 5.9 & \underline{94.7} $\pm$ 6.0 \\
    \bottomrule    
    \end{tabular}
    \label{tab:long-recording-patient-500}
\end{table*}
\begin{table*}[h!]
    \centering
    \caption{\textbf{Per patient comparison between {\densenetholter} and {\ikrnetholter}}. The models are trained on {\generepolholter} augmented with ECGs at 180, 215 and 500 Hz. The best accuracy is highlighted in \textbf{bold}, and the best F1 score is highlighted with \underline{underline}. Borderline cases are omitted, i.e., when the models predict only negative samples for a patient or when the patient contains only true negative samples.}
    \begin{tabular}{c|c|c|c|c||c|c|c|c}
    \toprule
       \multirow{3}{*}{\textbf{$F_s$}} & \multicolumn{4}{c||}{\textbf{\densenetholter}} & \multicolumn{4}{c}{\textbf{\ikrnetholter}} \\
        \cmidrule{2-9}
          & Accuracy  & Precision & Recall & F1 Score & Accuracy & Precision & Recall & F1 Score \\
          \midrule
          \multicolumn{9}{c}{\textbf{Test on {\generepol} dataset}} \\
          \midrule
150 & 94.3 $\pm$ 11.7 & 96.4 $\pm$ 13.1 & 92.8 $\pm$ 15.9 & 94.9 $\pm$ 10.3 & \textbf{97.3} $\pm$ 7.1 & 98.6 $\pm$ 5.7 & 97.0 $\pm$ 8.7 & \underline{97.4} $\pm$ 6.0 \\
180 & 94.6 $\pm$ 11.0 & 96.2 $\pm$ 13.1 & 93.6 $\pm$ 15.4 & 95.3 $\pm$ 9.8 & \textbf{97.9} $\pm$ 6.2 & 98.8 $\pm$ 5.6 & 97.8 $\pm$ 6.8 & \underline{98.1} $\pm$ 5.1 \\
215 & 95.2 $\pm$ 11.5 & 96.9 $\pm$ 12.8 & 93.5 $\pm$ 16.3 & 95.4 $\pm$ 10.7 & \textbf{96.0} $\pm$ 12.0 & 95.6 $\pm$ 16.4 & 98.0 $\pm$ 5.8 & \underline{97.3} $\pm$ 7.9 \\
250 & 94.6 $\pm$ 10.9 & 97.8 $\pm$ 8.0 & 92.4 $\pm$ 16.0 & 94.8 $\pm$ 10.1 & \textbf{98.2} $\pm$ 6.1 & 99.1 $\pm$ 5.2 & 98.0 $\pm$ 6.5 & \underline{98.3} $\pm$ 5.0 \\
300 & 95.2 $\pm$ 11.5 & 98.2 $\pm$ 8.6 & 92.8 $\pm$ 16.5 & 95.2 $\pm$ 11.1 & \textbf{98.0} $\pm$ 6.4 & 99.0 $\pm$ 5.3 & 97.6 $\pm$ 7.0 & \underline{98.1} $\pm$ 5.3 \\
350 & 95.0 $\pm$ 11.7 & 98.1 $\pm$ 8.7 & 92.7 $\pm$ 16.2 & 95.2 $\pm$ 10.9 & \textbf{98.1} $\pm$ 5.8 & 99.1 $\pm$ 4.8 & 97.6 $\pm$ 7.0 & \underline{98.2} $\pm$ 5.1 \\
425 & 94.9 $\pm$ 11.5 & 98.1 $\pm$ 8.4 & 92.8 $\pm$ 16.3 & 95.2 $\pm$ 10.7 & \textbf{98.1} $\pm$ 5.8 & 99.1 $\pm$ 4.8 & 97.6 $\pm$ 7.0 & \underline{98.2} $\pm$ 5.1 \\
500 & 94.9 $\pm$ 11.6 & 98.3 $\pm$ 8.4 & 92.6 $\pm$ 16.4 & 95.1 $\pm$ 10.8 & \textbf{98.1} $\pm$ 5.8 & 99.1 $\pm$ 4.8 & 97.6 $\pm$ 7.0 & \underline{98.2} $\pm$ 5.1 \\
          \midrule
          \multicolumn{9}{c}{\textbf{Test on {\generepolholter} dataset}} \\
          \midrule
150 & 89.8 $\pm$ 8.8 & 89.2 $\pm$ 10.9 & 92.6 $\pm$ 12.2 & 89.8 $\pm$ 10.2 & \textbf{92.6} $\pm$ 7.3 & 90.5 $\pm$ 9.1 & 96.8 $\pm$ 6.5 & \underline{93.1} $\pm$ 6.3 \\
180 & 90.6 $\pm$ 8.2 & 89.5 $\pm$ 10.5 & 94.0 $\pm$ 10.5 & 90.8 $\pm$ 8.6 & \textbf{93.2} $\pm$ 7.0 & 91.0 $\pm$ 9.0 & 97.3 $\pm$ 5.0 & \underline{93.7} $\pm$ 5.8 \\
215 & 90.7 $\pm$ 8.1 & 89.8 $\pm$ 10.1 & 93.7 $\pm$ 10.5 & 90.8 $\pm$ 8.4 & \textbf{92.6} $\pm$ 7.7 & 90.2 $\pm$ 9.7 & 97.5 $\pm$ 4.3 & \underline{93.3} $\pm$ 6.1 \\
250 & 91.4 $\pm$ 7.9 & 91.1 $\pm$ 9.8 & 93.5 $\pm$ 10.5 & 91.5 $\pm$ 8.2 & \textbf{93.8} $\pm$ 6.8 & 92.0 $\pm$ 8.7 & 97.3 $\pm$ 4.8 & \underline{94.2} $\pm$ 5.6 \\
300 & 92.0 $\pm$ 7.6 & 91.8 $\pm$ 9.4 & 93.7 $\pm$ 10.1 & 92.0 $\pm$ 8.0 & \textbf{94.0} $\pm$ 6.6 & 92.4 $\pm$ 8.6 & 97.2 $\pm$ 4.8 & \underline{94.4} $\pm$ 5.6 \\
350 & 91.9 $\pm$ 7.7 & 91.7 $\pm$ 9.4 & 93.6 $\pm$ 10.3 & 91.9 $\pm$ 8.2 & \textbf{94.1} $\pm$ 6.7 & 92.5 $\pm$ 8.5 & 97.2 $\pm$ 4.6 & \underline{94.5} $\pm$ 5.5 \\
425 & 92.1 $\pm$ 7.6 & 91.8 $\pm$ 9.4 & 93.8 $\pm$ 10.3 & 92.1 $\pm$ 8.3 & \textbf{94.1} $\pm$ 6.7 & 92.5 $\pm$ 8.5 & 97.3 $\pm$ 4.7 & \underline{94.5} $\pm$ 5.6 \\
500 & 92.0 $\pm$ 7.6 & 91.9 $\pm$ 9.4 & 93.6 $\pm$ 10.2 & 92.0 $\pm$ 8.2 & \textbf{94.1} $\pm$ 6.7 & 92.6 $\pm$ 8.5 & 97.1 $\pm$ 4.8 & \underline{94.5} $\pm$ 5.6 \\
    \bottomrule    
    \end{tabular}
    \label{tab:long-recording-patient-+}
\end{table*}  

\end{document}